\documentclass[sigconf]{acmart}
\AtBeginDocument{%
  }

\setcopyright{acmlicensed}
\copyrightyear{2025}
\acmYear{2025}
\acmDOI{10.1145/3746027.3755159}
\acmConference[MM '25]{Proceedings of the 33rd ACM International Conference on Multimedia}{October 27--31, 2025}{Dublin, Ireland}
\acmBooktitle{Proceedings of the 33rd ACM International Conference on Multimedia (MM '25), October 27--31, 2025, Dublin, Ireland}
\acmISBN{979-8-4007-2035-2/2025/10}

\usepackage{graphicx}
\usepackage{cases}
\usepackage{color}
\usepackage{microtype}
\usepackage{enumitem}
\usepackage{array,multirow,hhline, textcomp}
\usepackage{colortbl}
\usepackage{xcolor}
\usepackage{balance}
\usepackage[normalem]{ulem}
\usepackage{supertabular,booktabs}
\usepackage{makecell}
\usepackage{subcaption}
\usepackage{caption}
\usepackage{siunitx}
\usepackage[vlined,ruled,linesnumbered]{algorithm2e}
\usepackage{pifont}% http://ctan.org/pkg/pifont
\newcommand{\cmark}{\ding{51}}%
\newcommand{\xmark}{\ding{55}}%
\usepackage{arydshln}
\usepackage{bm}

\newcommand{\method}{\textit{PFDepth}}
\newcommand{\PrivateDatasetOne}{\textit{RealHet}}
\newcommand{\IdeaOne}{HSF}
\newcommand{\IdeaTwo}{3DGS}
\definecolor{darkgreen}{rgb}{0.0, 0.7, 0.0}

\definecolor{cvprblue}{rgb}{0.21,0.49,0.74}

\usepackage{soul}

\usepackage[capitalise]{cleveref}

\Crefname{figure}{Fig.}{Figs.}
\Crefname{table}{Tab.}{Tabs.}
\Crefname{section}{Sec.}{Secs.}
\Crefname{algorithm}{Alg.}{Algs.}

\begin{document}

\title{PFDepth: Heterogeneous Pinhole-Fisheye Joint Depth Estimation via Distortion-aware Gaussian-Splatted Volumetric Fusion}

\author{Zhiwei Zhang}
\authornote{Both authors contributed equally to this research.}
\email{zhangzw12319@sjtu.edu.cn}
\orcid{0009-0003-5488-8180}
\author{Ruikai Xu}
\orcid{0009-0003-9770-5938}
\authornotemark[1]
\email{xuruikai@sjtu.edu.cn}
\affiliation{%
  \institution{Shanghai Jiao Tong University,}
  \city{Shanghai}
  \country{China}
}

\author{Weijian Zhang}
\email{52265901038@stu.ecnu.edu.cn}
\orcid{0009-0006-5406-1011}
\author{Zhizhong Zhang}
\email{zzzhang@cs.ecnu.edu.cn}
\orcid{0000-0001-6905-4478}
\author{Xin Tan}
\email{xtan@cs.ecnu.edu.cn}
\orcid{0000-0001-9346-1196}
\affiliation{
  \institution{East China Normal University,}
  \city{Shanghai}
  \country{China}
}

\author{Jingyu Gong}
\authornote{Corresponding Authors.}
\email{jygong@cs.ecnu.edu.cn}
\orcid{0000-0002-4536-0953}
\affiliation{%
 \institution{East China Normal University,}
 \city{Shanghai}
 \country{China}
 }
 \affiliation{%
 \institution{Shanghai Key Laboratory of Computer Software Evaluating and Testing,}
 \city{Shanghai}
 \country{China}
 }

\author{Yuan Xie}
\email{yxie@cs.ecnu.edu.cn}
\orcid{0000-0001-6945-7437}
\affiliation{%
  \institution{East China Normal University,}
  \city{Shanghai}
  \country{China}
}

\author{Lizhuang Ma}
\authornotemark[2]
\email{lzma@sjtu.edu.cn}
\orcid{0000-0003-1653-4341}
\affiliation{%
  \institution{Shanghai Jiao Tong University,}
  \city{Shanghai}
  \country{China}
  }

\renewcommand{\shortauthors}{Zhiwei Zhang et al.}
%% No italics, no superscripts
%% Use footnote or author note to identify equal contribution and/or contact author info

\begin{abstract}
In this paper, we present \textbf{\textit{the first}} pinhole-fisheye framework for heterogeneous multi-view depth estimation, \method{}. Our key insight is to exploit the complementary characteristics of pinhole and fisheye imagery (undistorted vs. distorted, small vs. large FOV, far vs. near field) for joint optimization. \method{} employs a unified architecture capable of processing arbitrary combinations of pinhole and fisheye cameras with varied intrinsics and extrinsics. Within \method{}, we first explicitly lift 2D features from each heterogeneous view into a canonical 3D volumetric space. Then, a core module termed Heterogeneous Spatial Fusion is designed to process and fuse distortion-aware volumetric features across overlapping and non-overlapping regions. Additionally, we subtly reformulate the conventional voxel fusion into a novel 3D Gaussian representation, in which learnable latent Gaussian spheres dynamically adapt to local image textures for finer 3D aggregation. Finally, fused volume features are rendered into multi-view depth maps. Through extensive experiments, we demonstrate that \method{} sets a state-of-the-art performance on \textit{KITTI-360} and \PrivateDatasetOne{} datasets over current mainstream depth networks. To the best of our knowledge, this is the first systematic study of heterogeneous pinhole-fisheye depth estimation, offering both technical novelty and valuable empirical insights.
\end{abstract}


\begin{CCSXML}
<ccs2012>
   <concept>
       <concept_id>10010147.10010178.10010224.10010225.10010233</concept_id>
       <concept_desc>Computing methodologies~Vision for robotics</concept_desc>
       <concept_significance>300</concept_significance>
       </concept>
   <concept>
       <concept_id>10010147.10010178.10010224.10010225.10010227</concept_id>
       <concept_desc>Computing methodologies~Scene understanding</concept_desc>
       <concept_significance>500</concept_significance>
       </concept>
   <concept>
       <concept_id>10010147.10010178.10010224.10010226.10010239</concept_id>
       <concept_desc>Computing methodologies~3D imaging</concept_desc>
       <concept_significance>300</concept_significance>
       </concept>
 </ccs2012>
\end{CCSXML}

\ccsdesc[500]{Computing methodologies~Scene understanding}
\ccsdesc[300]{Computing methodologies~Vision for robotics}
\ccsdesc[300]{Computing methodologies~3D imaging}

\keywords{multi-view depth estimation, heterogeneous pinhole-fisheye cameras, volumetric spatial fusion, 3D Gaussian Splatting}

\begin{teaserfigure}
  \centering
  \vspace{-1em}
  \includegraphics[width=0.9
  \textwidth]{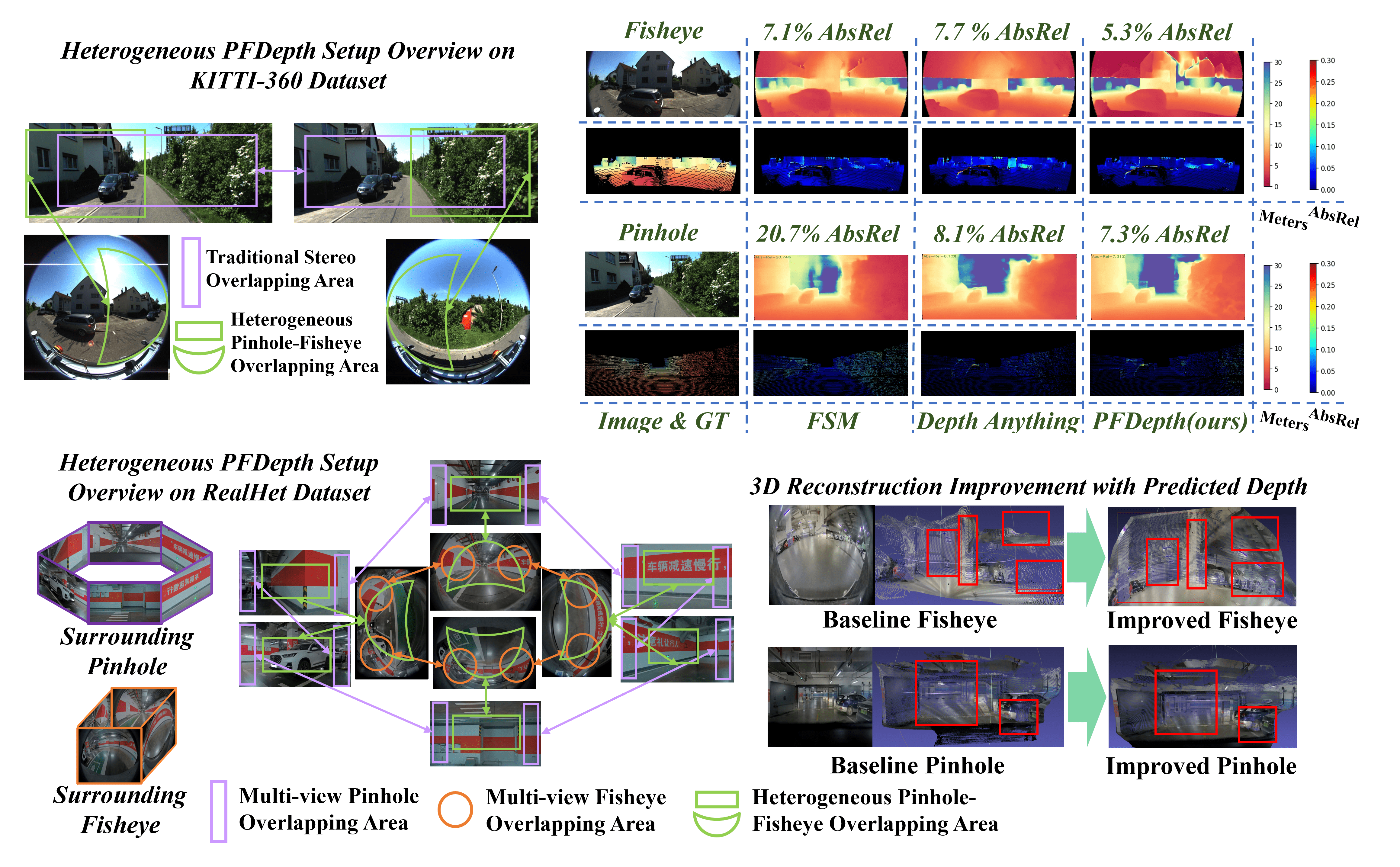}
   \vspace{-1em}
  % \vspace{-1em}
  \caption{We propose the first heterogeneous \textit{P}inhole-\textit{F}isheye \textit{Depth} Estimation network (\method{}) for joint optimization. By harnessing the complementary geometries of wide-FoV and narrow-FoV imagery, \method{} achieves notable accuracy gains in multi-view depth perception.}
  \Description{}
  \label{fig:teaser}
\end{teaserfigure}


\maketitle

%% Introduction
\section{Introduction}
\label{sec:intro}

Multi-view depth estimation is essential for autonomous driving and robotic navigation, particularly in vision-based systems. Accurate depth prediction ensures safe distance estimation between the ego-vehicle and surrounding obstacles. Although recovering depth from 2D images is an inherently ill-posed task, recent advances in both monocular and multi-view approaches have yielded significant progress. Notably, recent zero-shot monocular methods have leveraged large foundation models (\textit{e.g.} transformer-based \cite{ Bhat_2023, yin2023metric3d, yang2024depth, Li_2024_patchfusion, lin2024prompting_da} or diffusion-based \cite{Zhao_2023_vpd, ke2024repurposing, Tosi_2024, Fu_2024_geowizard, guizilini2024grin} architectures) with strong pre-trained weights to regress metric or relative depth (disparity) values, achieving fine-grained estimation performance and robust generalization ability. Concurrently, multi-view depth estimation has been further enhanced by designing specialized multi-view information aggregation modules and leveraging various spatial-temporal photometric self-supervision algorithms \cite{Yang_2021_selfmvs, guizilini2022full, kim2022self, Guizilini_2022_mfsdt, wei2023surrounddepth, Zhong_2023_multiframe, cheng2024adaptive}.

%---------------------------------------------------------------------
\begin{figure}[htbp!]
    \vspace{-1em}
    \centering
    \includegraphics[width=0.9\linewidth]{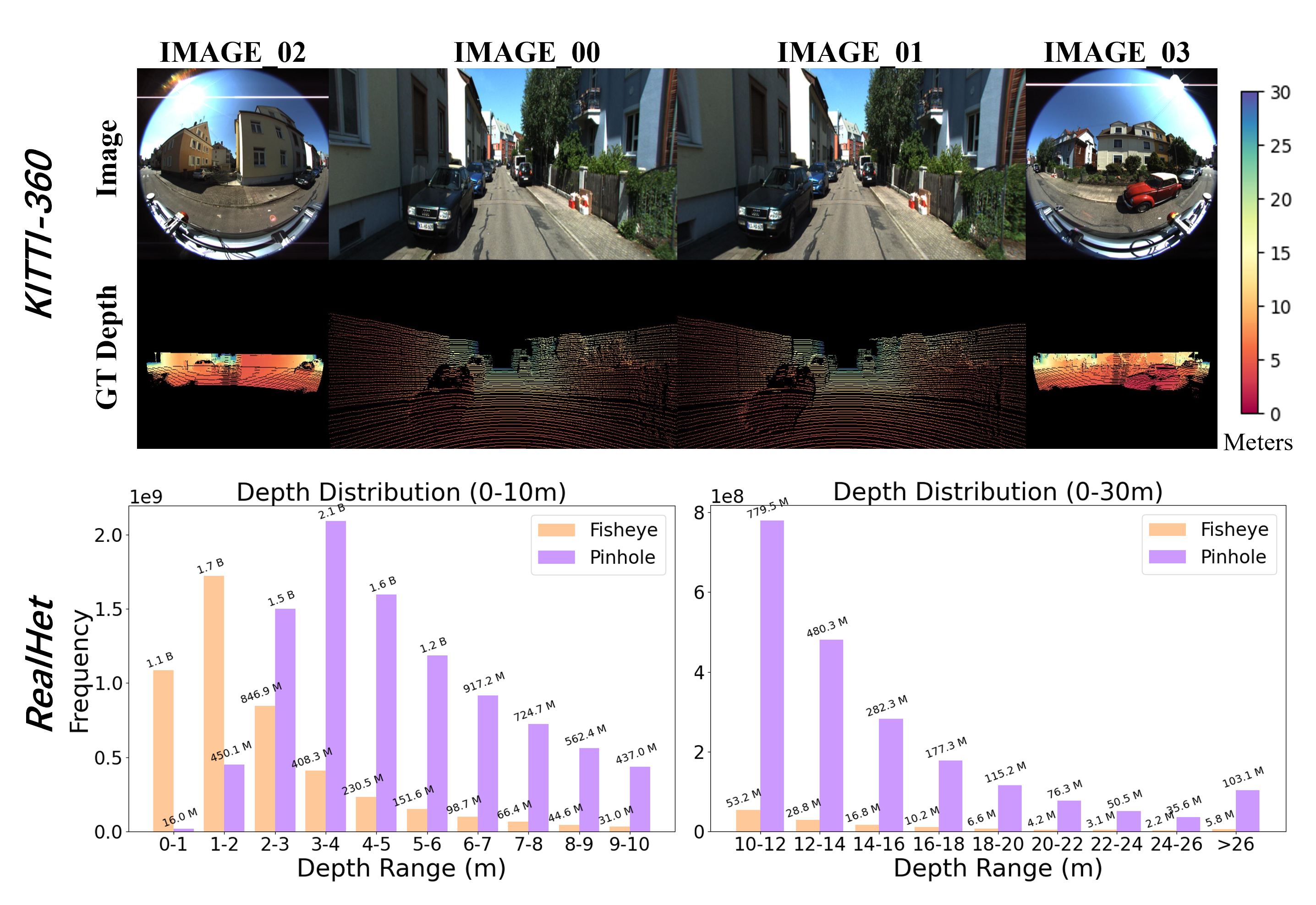}
    \vspace{-1.5em}
    \caption{Depth distribution comparison between pinhole and fisheye images (better zoomed-in).
    \Description{}
    }
    \vspace{-2em}
    \label{fig:intro_depth_dist}
\end{figure}

Despite recent great progress, issues and limitations are still encountered when state-of-the-art depth estimation networks are deployed in real-world environments. One of the key limitations arises from overlooking the heterogeneous device setup combining multiple pinhole and fisheye cameras. Pinhole-fisheye combination is widely utilized in autonomous vehicles and robots, because it offers unique advantages over homogeneous configurations as follows: 1) Pinhole and fisheye cameras capture distinct and complementary depth distribution patterns, which improve depth estimation covering near, side, and far-field regions (proved in \cref{fig:intro_depth_dist} and \cref{fig:exp_hsf_ablate}). 2) In contrast to pinhole-only systems, pinhole-fisheye cameras typically share larger overlapping areas, yielding denser cross-view correspondences and higher multi-view perception accuracy. 3) Compared to fisheye-only setups, pinhole-fisheye cameras are usually installed at different heights and poses in a heterogeneous manner, thus creating complex and asymmetric epipolar geometries that provide richer depth cues (detailedly visualized in \cref{fig:teaser}).

Yet, we have observed that current zero-shot monocular depth networks suffer from domain gaps and yield sub-optimal predictions on fisheye images, especially those with large FOV and severe visual distortion (demonstrated in \cref{table:quan_main} and \cref{fig:vis_compare_fisheye}). Moreover, re-training a fisheye-specific Depth Anything \cite{yang2024depth} model is rather difficult due to the shortage of open-source, high-quality, and densely annotated fisheye depth estimation datasets. Meanwhile, current multi-view depth estimation methods primarily target homogeneous setups, either pinhole-only \cite{kim2022self, wei2023surrounddepth} or fisheye-only \cite{kumar2021omnidet, xie2023omnividar}, leaving the great potential of heterogeneous pinhole-fisheye depth estimation unexplored.

%---------------------------------------------------------------------

Observing the underexplored mutual benefits of heterogeneous camera characteristics,
we find it necessary to propose a new proxy for multi-view metric depth estimation, termed \textbf{\textit{P}inhole-\textit{F}isheye \textit{Depth} estimation (\method{})}. Our motivation is to exploit the complementary visual information from both undistorted and distorted camera views, with the aim of optimizing them jointly.
Under this brand-new setting, we introduce our dedicated Pinhole-Fisheye Network, which can process arbitrary numbers of pinhole and fisheye images with varying intrinsic and extrinsic parameters and produce per-view depth estimations. \method{} employs calibrated projection and unprojection functions tailored for pinhole or multiple fisheye camera models to lift each 2D feature into 3D space. These features are then aggregated and encoded into a shared canonical volumetric representation, from which per-view depth is eventually rendered. 

%---------------------------------------------------------------------

The core component of \method{} is the Heterogeneous Spatial Fusion (HSF) module, which performs distortion-aware volume fusion in overlapping and non-overlapping regions among heterogeneous camera views. Additionally, we find that the pure voxel-based fusion method is coarse-grained and static, making it uneasy to capture both distorted and undistorted textures simultaneously. Drawing inspiration from multi-view 3D Gaussian research \cite{kerbl3Dgaussians, Charatan_2024_pixelsplat, Chen_2024_mvsplat, nips_MinLS024, Tian_2025_drivingforward, xu2024depthsplat}, we strategically extend the static voxel sampling to a dynamic Gaussian-Splatted sampling strategy. Specifically, essential 3DGS parameters, such as means, covariance matrices, and color features, are dynamically generated from our design of voxel-image cost volumes computation, allowing movable Gaussian spheres to actively and effectively align and match each heterogeneous view (\textit{e.g.} fitting with distinct visual distortion and camera geometry). Combining voxel-based and Gaussian-based fusion can further enhance the flexibility and robustness of depth inference in heterogeneous systems. 

%---------------------------------------------------------------------

Our contribution can be summarized as follows:
\begin{itemize}
    % \vspace{-1em}
    \item We propose the first \textbf{pinhole-fisheye heterogeneous depth estimation} network, \method{}. Under this novel setting, we design a unified framework that supports arbitrary camera configurations with diverse intrinsics and extrinsics, and jointly optimizes depth prediction for all views.

    \item We introduce a \textbf{Heterogeneous Spatial Fusion} module that explicitly models overlapping and non-overlapping regions across any pinhole-pinhole, pinhole-fisheye, and fisheye-fisheye configuration. 
    
    \item We propose \textbf{a 3DGS-based dynamic sampling and fusion strategy} as an extension to static voxel fusion. The learnable 3DGS spheres are dynamically generated and rendered, guided by distortion level, camera geometry, and image textures of each view, and thus largely enhancing estimation accuracy.

    \item Our approach achieves state-of-the-art performance on KITTI-360 and RealHet datasets. To our best knowledge, this is also \textbf{the first work} to systematically explore heterogeneous multi-view depth estimation with both technical and empirical insights.
\end{itemize}

%---------------------------------------------------------------------

%% Related Work
\section{Related Work}
\label{sec:related_work}

\textbf{Zero-shot Monocular Depth Estimation.} 
Currently, monocular depth estimation research has made great progress in universal zero-shot estimation across diverse datasets and scenes. Two main technical roadmaps have been proposed: the first approach involves constructing a discriminative foundation model using large vision transformers pre-trained on diverse datasets, while the second approach fine-tunes generative diffusion models to generate depth maps conditioned on input images.
Although transformer-based methods (\textit{e.g.}, ZoeDepth \cite{Bhat_2023},  Metric3D-v2 \cite{hu2024metric3d},  Depth Anything \cite{yang2024depth}, UniDepth \cite{piccinelli2024unidepth}) exhibit outstanding zero-shot performance on pinhole images, they underperform on fisheye images due to large visual distortion effects. Also, the scarcity of open-source, high-quality, densely annotated fisheye datasets hinders the training of a fisheye Depth Anything model.
On the other hand, while depth diffusion models (\textit{e.g.}, VPD \cite{Zhao_2023_vpd}, Marigold \cite{ke2024repurposing}, GeoWizard \cite{Fu_2024_geowizard}) can capture intricate geometric details with high-quality depthmap generation, their iterative denoising mechanism results in slow inference speed (5-65 seconds per 720P image, depending on varying denoising steps), hindering real-time application in autonomous system. 

Rather than merely enlarging a zero-shot pinhole model, which is hampered by scarce heterogeneous data and distortion-induced feature misalignment, we propose an innovative Pinhole-Fisheye framework in a multi-view manner to overcome these issues. Experiments show that leveraging the heterogeneous complementary information with static and dynamic samplings is more cost-effective than incorporating a large backbone network (See \cref{table:quan_main}). Our method also maintains faster inference speed than diffusion models.
%---------------------------------------------------------------------

\textbf{Multi-view Depth Estimation.} 
 Multi-view depth estimation networks can effectively leverage distinct visual information from multiple views to determine the most likely depth distribution, thereby benefiting 3D reconstruction. Early research, such as FSM \cite{guizilini2022full}, employs a shared encoder-decoder design and surround-view image aggregation to perform surrounding view depth estimation. Surround-Depth \cite{wei2023surrounddepth} utilizes a cross-view transformer for better multi-view feature aggregation, achieving superior performance. Currently, AFNet \cite{cheng2024adaptive} proposes a hybrid depth estimation model that leverages predictions from both monocular and multi-view depth networks to enhance dynamic object depth estimation. VFDepth \cite{kim2022self} is the closest to our methodology, as it adopts a regular 3D voxel and pinhole unprojection function to lift and aggregate homogeneous pinhole features. For comparison, our work stems from studying the more challenging heterogeneous fisheye-pinhole settings, supporting distortion-aware spatial fusion, and employing 3DGS as a novel representation for dynamic feature sampling. Our approach better bridges the gap of varying visual distortions and texture patterns across heterogeneous views.

%---------------------------------------------------------------------

\textbf{Fisheye Depth Estimation.}
Currently, there exist monocular fisheye depth networks, such as Fisheye-Distance-Net \cite{cheng2024adaptive} and SlaBins \cite{lee2023slabins} that directly train fisheye images from scratch in an end-to-end manner to learn visual distortion patterns; as well as surround-view fisheye network such as OmniDet \cite{kumar2021omnidet}, which combines fisheye depth estimation with other semantic/instance segmentation tasks, and OmniViDAR \cite{xie2023omnividar} that undistorts fisheye images into binocular pinhole images to train multi-view network. 
However, our method distinguishes itself by exploiting the complementary benefits of pinhole–fisheye systems: fisheye images excel in near-field perception, while pinhole images are more effective for medium-to far-distance perception. 
%---------------------------------------------------------------------

%% Method
\section{Methodology}
\label{sec:method}

%%%%%%%%%%%%%%%%%%%%%%%%%%%%% Problem Statement and Overall Formulation %%%%%%%%%%%%%%%%%%%%%%%%%%%%%
\subsection{Problem Statement}
\label{sec:pro_statement}
The heterogeneous multi-view depth estimation problem aims to infer pixel-level depth (or distance) simultaneously from surrounding pinhole and fisheye cameras to perceive a 360-degree environment. Specifically, given two sets of multi-view images at timestamp $t$, $\{\mathbf{I}_{P_i}^t\}_{i=1}^m$ captured by $m$ pinhole cameras, and $\{\mathbf{I}_{F_j}^t\}_{j=m+1}^{m+n}$ captured by $n$ fisheye cameras, we propose a unified heterogeneous depth network $\Phi$ to predict the depth map $\mathbf{Y}^t = \Phi\left\{ \mathbf{I}_{\{P_i, F_j\}}^t, \mathbf{E}_{\{i, j\}}, \mathbf{K}_{\{i, j\}}, \mathbf{\omega}_j \right\}$ for each view. Here, the subscript $P$ and $F$ are abbreviations of Pinhole and Fisheye images, respectively. Additionally, the corresponding camera extrinsic matrix $\mathbf{E}$, intrinsic matrix $\mathbf{K}$ and distortion coefficients $\mathbf{\omega}$ (specific to fisheye camera models) are also assumed to be known as inputs. Our heterogeneous framework can accommodate any combination of $m+n$ pinhole and fisheye cameras, with arbitrary camera poses.

%%%%%%%%%%%%%%%%%%%%%%%%%%%%% Network Architecture %%%%%%%%%%%%%%%%%%%%%%%%%%%%%
\subsection{Network Architecture}
\label{sec:network_archi}

\begin{figure*}[t!]
    \centering
    \includegraphics[width=\linewidth]{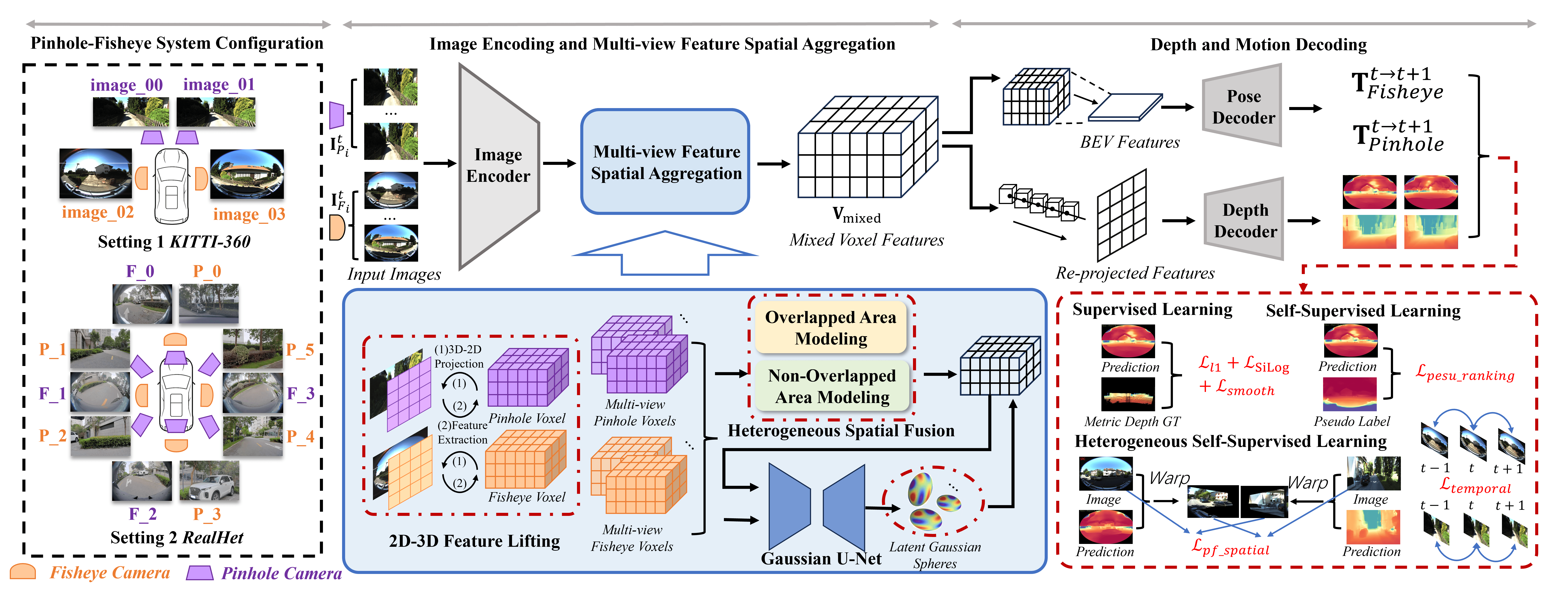}
    \vspace{-1em}
    \caption{The overall framework of \method{}. The input images are first processed through image encoding, followed by multi-view spatial feature aggregation and subsequent depth and motion decoding. The core of our multi-view aggregation lies in the Heterogeneous Spatial Fusion module, which integrates 2D-to-3D feature lifting, explicit modeling of overlapped and non-overlapped regions, and a 3D Gaussian Spheres (3DGS) enhancement for dynamic sampling and fusion operations into a mixed voxel.
    }
    \Description{}
    \vspace{-2em}
    \label{fig:method_main_network}
\end{figure*}

 The whole structure of \method{} is illustrated in \cref{fig:method_main_network}. It begins by encoding multi-view images ($\mathbf{I}_P$ and $\mathbf{I}_F$) with a shared 
 encoder to produce feature maps ($\tilde{\mathbf{I}}_P$ and $\tilde{\mathbf{I}}_F$). Subsequently, the multi-view feature spatial aggregation process (\cref{sec:msa}) follows, which consists of a projection-based distortion-aware feature lifting method to transform 2D features to 3D voxel features, and then an HSF module to determine and combine overlapped and non-overlapped regions across views into a unified volumetric representation capturing diverse spatial information (static aggregation period). Next, a modified 3D Gaussian Splatting process (\cref{sec:3dgs}) dynamically samples and aggregates surrounding perspective features by estimating per-pixel movable Gaussian spheres from image features, cost volumes, and voxel features. The former static voxel features are re-projected using a NeRF-like renderer and consecutively concatenated with the pixel-wise Gaussian features. A depth decoder then predicts disparity maps, and meanwhile, voxel features across frames are pooled into Bird’s Eye View (BEV) for camera pose estimation (\cref{sec:depth_motion_decode}). Finally, the network is trained end-to-end using supervised and self-supervised losses (\cref{sec:training_loss}).

%%%%%%%%%%%%%%%%%%%%%%%%%%%%% Multi-view Feature Spatial Aggregation(MSA) %%%%%%%%%%%%%%%%%%%%%%%%%%%%%
\subsection{Multi-view Feature Spatial Aggregation}
\label{sec:msa}

\noindent \textbf{ Distortion-aware Feature Lifting Method.}
In previous research, various methods for homogeneous multi-view feature fusion have been explored, such as shared encoder-decoder architectures \cite{kumar2021omnidet, guizilini2022full}, cross-view attention \cite{wei2023surrounddepth, xie2023omnividar}, and unified BEV space techniques \cite{mao2023bevscope}, among others. However, the pinhole-fisheye configuration presents unique challenges for aforementioned approach due to large differences varied in distortion, FOV, and lighting conditions across heterogeneous views. Motivated by VFDepth \cite{kim2022self}, we introduce distortion-aware camera models to first undistort features and elevate them into a unified canonical 3D space via a projection-based manner.

% --- 2D-3D Lifting Formula
\begin{align}
    \label{formula:2d_3d_lifting}
    \mathbf{s}' &= \Pi_P (\mathbf{s}, \mathbf{E}_i, \mathbf{K}_i), \\
    \mathbf{V}_i[\mathbf{s}] &= \text{Bilinear-Interpolate} (\tilde{\mathbf{I}}_{P_i}, \mathbf{s}'), \\
    \mathbf{s}'' &= \Pi_F (\mathbf{s}, \mathbf{E}_j, \mathbf{K}_j, \omega_j), \\
    \mathbf{V}_j[\mathbf{s}] &= \text{Bilinear-Interpolate} (\tilde{\mathbf{I}}_{F_j}, \mathbf{s}''), 
\end{align}

\noindent where $\mathbf{V}_i[\mathbf{s}]$ and $\mathbf{V}_j[\mathbf{s}]$ represent $i$ th pinhole camera's and  $j$ th fisheye camera's voxel features at the voxel center position $\mathbf{s}$, while $\mathbf{s}', \mathbf{s}'' = [u,v,1]^{\text{T}}$ represent the projected locations in images for each camera type.

For pinhole model, the projection function $\Pi_P$ and its back-projection function $\Pi_P^{-1}$ are formulated as follows:

% --- pinhole proj_unproj
\begin{align}
    \label{formula:pinhole_proj}
    \Pi_P(\mathbf{s}, \mathbf{E}_i, \mathbf{K}_i) &= \frac{1}{d} \mathbf{K}_i \mathbf{E}_i \mathbf{s}, \\
    \label{formula:pinhole_unproj}
    \Pi_P^{-1}(\mathbf{s}', \mathbf{E}_i, \mathbf{K}_i, d) &= d \mathbf{E}_i^{-1} \mathbf{K}_i^{-1} \mathbf{s}'.
\end{align}

\noindent where the depth value, denoted as $d$, is obtained during the projection process from the third entry of $\mathbf{K_i}\mathbf{E_i}\mathbf{s}$. Conversely, during the back-projection process, $d$ is derived from a predefined depth bin value (Used in depth decoding of \cref{sec:depth_motion_decode}).

For fisheye models, we incorporate two distinct fisheye camera models to enhance generality. For the \PrivateDatasetOne{} that utilizes Kannala-Brandt models (KB model) \cite{Kannala_2006}, we define fisheye projection function 
$$\Pi_F(\mathbf{s}, \mathbf{E}_j, \mathbf{K}_j, \omega_j)$$ 
\noindent as follows:

% --- fisheye proj
\begin{align}
        \label{formula:proj_kb_fisheye_1}
     &\left[  \begin{matrix}
         x_c & y_c & d & 1 \\
     \end{matrix} \right]^\text{T} 
      = \mathbf{E}_j \mathbf{s}, \\
        \label{formula:proj_kb_fisheye_2}
        &\left[  \begin{matrix}
         a & b & 1 & 1 \\
     \end{matrix} \right]^\text{T}  = \frac{1}{d} \left[  \begin{matrix}
         x_c & y_c & d & 1 \\
     \end{matrix} \right]^\text{T}, \\
       \label{formula:proj_kb_fisheye_3}
      &r(\theta) = \sqrt{a^2 + b^2} = \frac{\sqrt{x_c^2 + y_c^2}}{d}= \tan(\theta), \\
       \label{formula:proj_kb_fisheye_4}
      &\phi(\theta) = \theta + \omega_j^1 \theta^3 + \omega_j^2 \theta^5 + \omega_j^3 \theta^7 + \omega_j^4 \theta^9, \\
      \label{formula:proj_kb_fisheye_5}
      &\mathbf{s}'' = \mathbf{K}_j \left[ \begin{matrix} \frac{\phi(\theta)}{r(\theta)} \cdot a & \frac{\phi(\theta)}{r(\theta)} \cdot b & 1 \end{matrix} \right ]^{\text{T}},
\end{align}
\noindent where $\theta$ is the incident angle of the input ray. As for the back-project function 
$$\Pi_F^{-1}(\mathbf{s}'', \mathbf{E}_j, \mathbf{K}_j, \omega_j, d),$$
\noindent it is derived as follows:

% --- fisheye unproj
\begin{align}
    \label{formula:back_kb_fisheye_1}
   & \left[ \begin{matrix} a^{'} & b^{'} & 1 \\ \end{matrix} \right ]^{\text{T}} = \mathbf{K}_j^{-1} \mathbf{s}^{''}, \\
    \label{formula:back_kb_fisheye_2}
   & \phi(\theta) = \sqrt{a^{'2} + b^{'2}}, \\
   \label{formula:back_kb_fisheye_3}
   & \theta = \phi^{-1} (\sqrt{a^{'2} + b^{'2}}), \\
   \label{formula:back_kb_fisheye_4}
   & r(\theta) = \tan(\theta), \\
   \label{formula:back_kb_fisheye_5}
   & \mathbf{s} = \mathbf{E}_j^{-1} \left[\begin{matrix} \frac{r(\theta)}{\phi(\theta)} \cdot a^{'} \cdot d & \frac{r(\theta)}{\phi(\theta)} \cdot b^{'} \cdot d & d & 1 \end{matrix} \right ]^{\text{T}},
\end{align}

\noindent where $\phi^{-1}$ represents the root of a univariate high-degree polynomial equation (up to nine). The root of the solution can be obtained by numerical approximation methods to pre-compute and saved down.

As for \textit{KITTI-360}, the dataset employs a simplified version of the MEI camera model \cite{Mei_2007} in implementation. Compared with the KB model, the only difference in the projection function $\Pi_F$ lies in the implementation of $\phi(\theta)$ and $r(\theta)$:

% --- MEI model
\begin{align}
& \textit{norm} = ||\mathbf{E}_j\mathbf{s}||_2, \\
& r(\theta) = \frac{\sqrt{x_c^2 + y_c^2}}{d + \epsilon \cdot \textit{norm}},\\
& \phi(\theta) = 1 + \omega_j^1 r(\theta)^2 + \omega_j^2 r(\theta)^4,
\end{align}

\noindent where the new camera parameter $\epsilon$ is denoted as mirror factor. The derived solution of the root equals that of the following equations:

\begin{align}
    \label{formula:eq_solve_r}
    (1-\sqrt{a^{'2} + b^{'2}}) + \omega_j^1 r^2 + \omega_j^2 r^4 &= 0, \\
    \label{formula:eq_solve_theta}
    \sin \theta - r \cos \theta - r \epsilon &= 0,
\end{align}
\noindent where $r$ is first solved from \cref{formula:eq_solve_r} and then deployed into \cref{formula:eq_solve_theta} to solve $\theta$.

 Meanwhile, the solution to the equation $\theta = \phi^{-1}(\sqrt{a^{'2} + b^{'2}})$ depends solely on camera parameters, and, as such, is pre-computed and stored in a lookup table prior to the training or inference period.

% ----------------------------------

\begin{figure}[htbp!]
    \centering
    \includegraphics[width=1.0\linewidth]{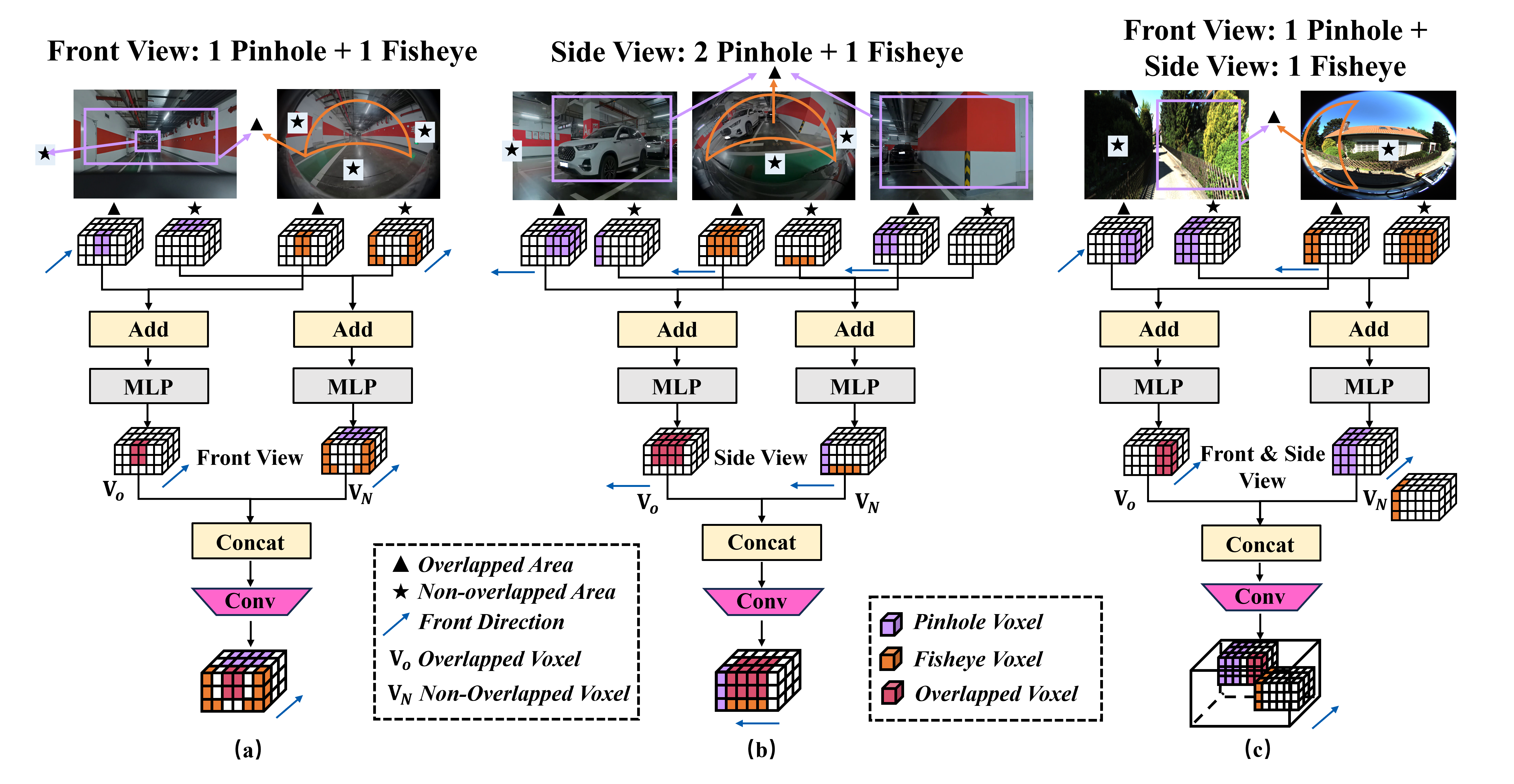}
    \vspace{-1em}
    \caption{ Overview of overlapped and non-overlapped area modeling in the Heterogeneous Spatial Fusion module.
    }
    \Description{}
    \label{fig:method_HSF}
    \vspace{-2em}
\end{figure}

\noindent \textbf{Heterogeneous Spatial Fusion Module.}
We present three typical configurations to illustrate the whole process in \cref{fig:method_HSF}: (a) one pinhole camera plus one fisheye camera in the front view (from \textit{\PrivateDatasetOne{}} Dataset), (b) two pinhole cameras with one fisheye camera in the right-side view (from \textit{\PrivateDatasetOne{}} Dataset), and (c) one pinhole camera in the front view plus one fisheye camera in the right-side view (from \textit{KITTI-360}). These examples encompass the majority of vision sensor arrangements. Our approach begins by identifying overlapping and non-overlapping voxel sets for each view-specific volumetric space, followed by the concatenation and fusion of the overlapping regions into a unified voxel space $\mathbf{V}_O$, and the aggregation of non-overlapping voxels into volumetric space $\mathbf{V}_N$. Finally, we combine$\mathbf{V}_O$ and $\mathbf{V}_N$ to form a comprehensive voxel representation, where dense heterogeneous features from different views and sensors are spatially interconnected.

% --- HSF figure

The determination of overlapped and non-overlapped voxels, as illustrated in \cref{alg:overlap_mask} in supplementary material, involves checking whether the projection location of a voxel falls onto more than one view plane. A voxel is categorized as overlapping between views $i$ and $j$ if both $\mathbf{Mask}[i]$ and $\mathbf{Mask}[j]$ are \textit{True}. As \cref{alg:hsf} in the supplementary material demonstrates, we then filter out overlapped and non-overlapped voxels between every two neighboring views, aggregate them respectively, and project into another latent feature space using two independent MLPs. For overlapped voxels, multi-view information from different sensors is exchanged and integrated at the same spatial position. For non-overlapped ones, complementary information from these sensors is accumulated together, as they do not share spatial positions. In our implementation, we iteratively traverse each view and perform the aforementioned operations to obtain $\mathbf{V}_O$ and $\mathbf{V}_N$. Ultimately, the two accumulated voxels are concatenated and transformed into mixed voxel features $\mathbf{V}_{\text{mix}}$ via convolution operations.

%%%%%%%%%%%%%%%%%%%%%%%%%%%%% 3DGS Enhancement %%%%%%%%%%%%%%%%%%%%%%%%%%%%%
\subsection{3DGS Dynamic Sampling Enhancement}
\label{sec:3dgs}

In this section, we propose to utilize \textbf{the dynamic alignment nature} of Gaussian spheres to enhance static multi-view voxel representation for improved depth prediction. Voxel-based fusion strategy adopts predetermined 3D positions for projection or unprojection, \textbf{establishing a fixed mapping between the pixels and static voxels.} However, limitations in static voxel representation arise from predefined positions and coarse voxel resolutions, which make it challenging to capture deformable local textures (e.g., marginal areas with significant distortion).  Although 3D Gaussian representation has a close connection with depth estimation, recent 3DGS methods \cite{kerbl3Dgaussians, Charatan_2024_pixelsplat, Chen_2024_mvsplat, nips_MinLS024, Tian_2025_drivingforward, xu2024depthsplat} primarily focus on improving novel view synthesis (NVS), while treating depth estimation as a side objective. In contrast, our proposed \method{} introduces movable Gaussian spheres as a dynamic sampling medium to capture 2D latent features as a complement for voxel features.

\begin{figure}[htbp!]
    \centering
    \includegraphics[width=1.0\linewidth]{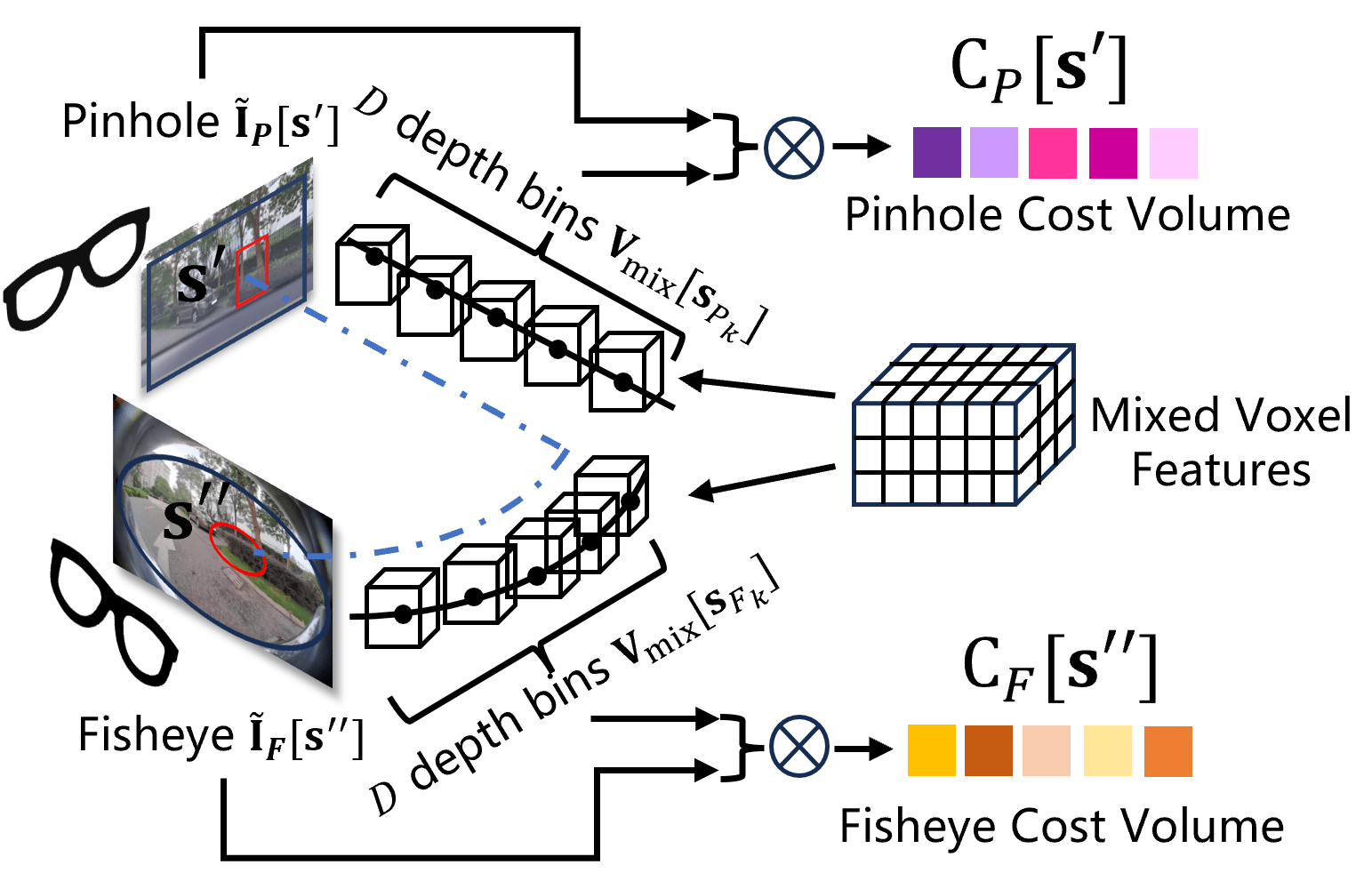}
    \caption{Illustration of pixel-voxel cost-volume calculation.
    }
    \Description{}
    \label{fig:method_costvolume}
    \vspace{-1em}
\end{figure}

Firstly, we construct the cost volume via calculating the similarity of pixels and the fused HSF voxel (shown in \cref{fig:method_costvolume}), different from MVSplat \cite{Chen_2024_mvsplat} and DepthSplat \cite{xu2024depthsplat}, which compute pair-wise cross-view warping as cost-volumes. We use re-projection strategy shown in \cref{formula:pinhole_back_proj}-\cref{formula:fisheye_back_proj}. For each pixel position in the pinhole (denoted as $\mathbf{s}'$) or fisheye image coordinates (denoted as $\mathbf{s}''$), where $\mathbf{s}', \mathbf{s}'' = [u, v, 1]^{\text{T}}$, we pre-define a set of fixed-size depth bins $ \hat{\mathbf{d}} = \{ \hat{d}_k \in \mathbb{R} \}_{k=1}^D $ to serve as $D$ sampling bins. Given back-project function in \cref{formula:pinhole_unproj} and \cref{formula:back_kb_fisheye_1}-\cref{formula:back_kb_fisheye_5} , we have

% --- formula: coarse depth decoding
\begin{align}
    \label{formula:pinhole_back_proj}
    \mathbf{s}_{P_k} &= \Pi_P^{-1}(\mathbf{s}', \mathbf{E}_i, \mathbf{K}_i, \hat{d}_k), \\
    \label{formula:fisheye_back_proj}
    \mathbf{s}_{F_k} &= \Pi_F^{-1}(\mathbf{s}'', \mathbf{E}_j, \mathbf{K}_j, \omega_j, \hat{d}_k), \\
    \label{formula:pinhole_costvolume}
    \mathbf{C}_P[\mathbf{s}'] &= \left\{ \frac{ \mathbf{V}_\text{mix}[\mathbf{s}_{P_k}] \cdot \mathbf{\tilde{I}}_P[\mathbf{s}']}{\sqrt{Channels}}) \right \}_{k=1}^D, \\
    \label{formula:fisheye_costvolume}
    \mathbf{C}_F[\mathbf{s}''] &= \left\{ \frac{\mathbf{V}_\text{mix}[\mathbf{s}_{F_k}] \cdot \mathbf{\tilde{I}}_F[\mathbf{s}'']}{\sqrt{Channels}} \right\}_{k=1}^D,
\end{align}

\noindent where $\mathbf{s}_{P_k}, \mathbf{s}_{F_k} \in \mathbb{R}^3 $ are back project 3D positions from pinhole and fisheye views given depth bin value $\hat{d}_k$.
Meanwhile, $\mathbf{C}_{ \{P, F\}} \in \mathbb{R}^{H \times W \times D}$ are cost-volume tensors, and $\mathbf{C}_{ \{P, F\}}[\cdot]$ indicates that each fisheye or pinhole pixel possesses $D$ similarity values with $D$ back-projected voxels from $\mathbf{V}_\text{mix}$. In practice, $\mathbf{V}_{\text{mix}}[\cdot]$ adopts triplet-interpolation operation to gather features at 3D position, and $\mathbf{\tilde{I}}_{\{P, F\}}[\cdot]$ uses bicubic-interpolation to gather features at 2D location. Then a weighted averaged calculation is adopted to obtain coarse center depth.

% --- formula: coarse-grained depth decoding
\begin{align}
    \label{formula:pinhole_weighted_average}
    \tilde{\mathbf{d}}_P[\mathbf{s}'] &= \text{Softmax}(\text{MLP}(\mathbf{C}_P[\mathbf{s}'])) \cdot \hat{\mathbf{d}}, \\
    \label{formula:fisheye_weighted_average}
    \tilde{\mathbf{d}}_F[\mathbf{s}''] &= \text{Softmax}(\text{MLP}(\mathbf{C}_F[\mathbf{s}''])) \cdot \hat{\mathbf{d}}, 
\end{align}

\noindent where $\tilde{\mathbf{d}}_{\{P, F\}} \in \mathbb{R}^{H \times W \times 1}$ represents the inferred depth of Gaussian sphere center at pixel $\mathbf{s}'$ or $\mathbf{s}''$. The 3D position of sphere center can be similarly calculated by averaging $\mathbf{s}_{P_k}$ or $\mathbf{s}_{F_k}$. We choose pixel-voxel cost volume instead of cross-view pixel-pixel warping to determine Gaussian centers, since heterogeneous views often share irregular overlapping areas in a 2D perspective view. The adoption of fused voxel features $\mathbf{V}_\text{mix}$ can instead provide a more robust and steady fused features to enhance the quality of cost-volume.

Finally, we design a U-Net with a residual structure to respectively regress the variance and sphere harmonics of color as follows (for both fisheye and pinhole views),

\begin{align}
    \label{formula:refined_fisheye_density}
    \mathbf{\rho} &= \text{max}_{\hat{d}} \left\{ \text{Softmax}\{\text{MLP}(\mathbf{C})\} \right\}, \\
    \label{formula:latenct_gaussian_feature}
    \mathbf{F}_\text{GS} &= \text{Residual-Unet}(\mathbf{\tilde{I}} \oplus \mathbf{C} \oplus \mathbf{\rho} \oplus \tilde{\mathbf{d}}),
\end{align}

\noindent where the output $\mathbf{F}_\text{GS}$ is composed of covariance matrix and color harmonics. The final concatenated Gaussian feature map $\tilde{\mathbf{d}} \oplus \mathbf{F}_\text{GS}$ will be interpolated to match the resolution of the last ResNet Encoder feature, so that it can be concatenated with the rendered feature from the fused voxel.

%%%%%%%%%%%%%%%%%%%%%%%%%%%%% Depth and Motion Decoding %%%%%%%%%%%%%%%%%%%%%%%%%%%%%
\subsection{Depth and Motion Decoding}
\label{sec:depth_motion_decode}

We follow \cref{sec:3dgs} and utilize the back-projection function \cref{formula:pinhole_back_proj} - \cref{formula:fisheye_back_proj} to elevate 2D coordinates into 3D positions for feature interpolation from $\mathbf{V}_{\text{mix}}$. Each pixel also aligns with a set of $D$ voxel features along its casting ray. Then, we employ an average pooling operation to compress multiple voxel features to one vector per pixel. Finally, we utilize a shared depth decoder to decode the disparity value from each view. Along with depth decoding, we also need pose prediction for the self-supervised learning protocol as \cite{guizilini2022full, kim2022self, wei2023surrounddepth}. Given mixed volumetric features $\mathbf{V}_{\text{mix}}$, we use 3D convolutions to reduce channels in height dimension and obtain BEV features at time $t$, and use light-weight decoder to regress $4 \times 4$ pose matrix.

%%%%%%%%%%%%%%%%%%%%%%%%%%%%% Training Loss and Strategies %%%%%%%%%%%%%%%%%%%%%%%%%%%%%
\subsection{Training Loss and Strategies}
\label{sec:training_loss}
In alignment with previous studies \cite{guizilini2022full, kim2022self, wei2023surrounddepth}, we adopt a combination of supervised and self-supervised strategies to train our innovative heterogeneous settings.
For supervised loss, we utilize a combination of L1 loss, SILog loss \cite{eigen2014depth}, and pseudo ranking loss \cite{chen2016single, xian2020structure} for ground-truth disparity supervision.
With the movement of the ego-vehicle, self-supervision can be performed utilizing the optical flow consistency between consecutive frames. We employ spatial and temporal image reconstruction loss \cite{kim2022self, Tan_2025} for self-supervised learning. We also import Edge-aware Smoothness Loss \cite{Xian_2020} to enforce boundary sharpness. The specific form of the loss function is provided in the supplementary materials.

%% Experiments
\section{Experiments}
\label{sec:experiments}

%%%%%%%%%%%%%%%%%%%%%%%%%%%%% Datasets %%%%%%%%%%%%%%%%%%%%%%%%%%%%%
\subsection{Datasets}
\label{sec:datasets}

% ----------------- Quantitative comparison -----------------
\begin{table*}[t]
    \centering
    \caption{
        \textbf{Quantitative comparison}
        of \method{} with SOTA monocular and multi-view depth estimators on \textit{fisheye} images evaluation under the heterogeneous fisheye-pinhole setting. 
        AbsRel and $\delta_1$ metrics$^\dagger$ are presented in percentage terms, while RMSE are presented in real number format. \textbf{Bold} numbers are the best, \underline{underscored} second best. GPU memory and FPS are tested on a single Tesla V100 32 GB with batch-size set as 1 during the inference period. 
    }
    \scriptsize
    \resizebox{\linewidth}{!}{
	\begin{tabular}{
% r, c, l means aligned to right, center and left. @{} means delete default column width, and the following p{2.0em} means setting fixed column width while p refers to `parabox`.
@{}l @{}c @{}c 
% c@{} c@{}p{1.0em}@{}
c@{\hspace{0.5em}} c@{\hspace{0.5em}} c@{\hspace{0.5em}} p{1.0em}@{}
c@{\hspace{0.5em}} c@{\hspace{0.5em}} c@{\hspace{0.5em}} p{1.0em}@{}
c@{\hspace{0.5em}} c@{\hspace{0.5em}} c@{\hspace{0.5em}}
}

\toprule

\multirow{2}{*}{Method} &
\multirow{2}{*}{Configuration} & 
\multirow{2}{*}{Backbone/Params} &
\multicolumn{3}{c}{\textit{KITTI-360} FOV$\approx \SI{220}{\degree} $} & &
\multicolumn{3}{c}{\PrivateDatasetOne{} FOV=$\approx \SI{160}{\degree} $} & &
\multicolumn{1}{c}{Memory} & &
\multicolumn{1}{c}{FPS}
\\

& & &
AbsRel↓ &
RMSE↓ & 
$\delta$1↑ & &
AbsRel↓ &
RMSE↓ &
$\delta$1↑ & &
(MB) &  & (Hz)
\\

\midrule

FSM (RA-L 2020)~\cite{guizilini2022full} & Multi-view & ResNet-50 (58M) &
23.2 & 3.025 & 74.4 & & 
7.3  & 0.527 & 94.3 & & 
450 & & 120.0
\\

VFDepth (NeuralIPS 2022)~\cite{kim2022self} & Multi-view & ResNet-50 (73M) &
9.1 & 1.878 & 90.6 & & 
6.9 & 0.543 & 94.6 & & 
1664 & & 73.2
\\

SurroundDepth (CoRL 2023)~\cite{wei2023surrounddepth} & Multi-view & ResNet-50 + CVT (78M) &
9.0 & 1.848 & 91.0 & & 
- &  - & - & & 
14336 & & 54.8
\\

ZoeDepth (ArXiv 2023)~\cite{Bhat_2023} & Monocular & BEiT384-L (345M) &
12.6 & 2.304 & 82.9 & & 
  -   &  -   &  -   & &
- & & -
\\
Depth Anything (CVPR 2024)~\cite{yang2024depth} & Monocular & ViT-L (335M) &
11.7 & 2.012 & 90.7 & & 
  -   &  -   &   -   & &
4537 & & 10 
\\

Marigold (CVPR 2024)$^*$~\cite{ke2024repurposing} &  Monocular & Stable-Diffusion V2 (890 M) &
15.6  & 2.620 &  88.4 & & 
28.2  & 2.274 &  54.5 & &
9625 & & 0.2
\\

\midrule

PFDepth-CE (Cost-Effective) & Heterogeneous & Res50+GS-UNet (130.8M) &
\underline{8.3} & \underline{1.714} & \underline{92.0}  & &
\textbf{6.0} & \textbf{0.524} & \textbf{95.9} & &
1946 & & 51.0 \\

PFDepth-BP (Best Performance) & Heterogeneous & Res50+ViT-L+GS-UNet (435M) &
\textbf{8.1} & \textbf{1.654} & \textbf{92.4}  & &
- & - & -  & &
2816 & & 9.1
\\

\bottomrule

\end{tabular}
	\label{table:quan_main}
    }
    \\
    \begin{minipage}{0.96\linewidth}
        \scriptsize
        \vspace{0.4em}
        \begin{itemize}
        \item[$^\dagger$]
        Since \textit{KITTI-360} hasn't released the official depth ground-truth, all of the baselines in our benchmark are trained by ourselves, using the same LiDAR-generated depth ground-truths under the same evaluation split, without any ensemble strategy. For zero-shot monocular SOTA methods. We fine-tune these models based on the officially released pre-trained weights.
        
        \item[$^*$] Due to VAE limitation, Marigold can only be trained or fine-tuned using dense depth labels. However, our dataset has only sparse LiDAR-projection depth maps for autonomous driving. Therefore, Marigold derives we adopts least squares fitting method provided by original paper to transform affine-invariant depth to metric depth for performance comparisons.
        \end{itemize}
    \end{minipage}
\end{table*}

% ----------------- Ablation Study of Heterogeneous Pinhole-Fisheye Configuration on KITTI-360  -----------------
\begin{table}[t]
    \centering
    \caption{
        \textbf{Ablation study}
        of the effectiveness in heterogeneous pinhole-fisheye configurations on \textit{KITTI-360} dataset. ``image\_00'' and ``image\_01'' refer to the two front pinhole cameras, while ``image\_02'' and ``image\_03'' represent the left and right fisheye cameras.
    }
    \resizebox{\linewidth}{!}{
	\large 
\begin{tabular}{c @{}cccc cc c@{}c c@{}c}
\toprule

\multicolumn{1}{c}{\multirow{2}{*}{Exp.}} &
\multicolumn{2}{c}{Pinhole Camera} &
\multicolumn{2}{c}{Fisheye Camera} &
\multicolumn{1}{c}{\multirow{2}{*}{\IdeaOne{}}} &
\multicolumn{1}{c}{\multirow{2}{*}{\IdeaTwo{}}} &
\multicolumn{2}{c}{Fisheye Performance} &
\multicolumn{2}{c}{Pinhole Performance}
\\
&
\multicolumn{1}{c}{image\_00} &
\multicolumn{1}{c}{image\_01} &
\multicolumn{1}{c}{image\_02} &
\multicolumn{1}{c}{image\_03} &
&
&
AbsRel↓ & 
$\delta$1↑ &
AbsRel↓ & 
$\delta$1↑
\\

\midrule      

(a) &\cmark &  &  &  &
\xmark & - &- & - & 8.4 & 90.7 \\

(b) & & & \cmark&  &
\xmark & - & 10.9 & 87.4 & - & - \\ 
 
(c) & \cmark & & \cmark&  &
\xmark & - & 12.7 & 85.0 & 28.6 & 41.1 \\ 

(d) & \cmark & & \cmark&  &
\cmark & - & 10.7 & 88.0 & 9.5 & 89.6 \\ 

\midrule
(e) & \cmark & \cmark &  &  &
\xmark & - & - &- & 9.4& 89.6\\

(f) & \cmark & \cmark &  &  &
\cmark & - & -&- & 7.9 & 91.4 \\

(g) & &   & \cmark & \cmark &
\xmark & - & 9.5 & 89.8 & - & - \\

(h)  &  &   & \cmark & \cmark &
 \cmark & - & 9.1 & 89.5 & - & -\\

(i) & \cmark &  \cmark & \cmark & \cmark &
\xmark & \xmark & 9.7 & 90.2 & 28.4 & 38.7 \\

(j) & \cmark   &  \cmark & \cmark & \cmark &
\cmark &\xmark & 8.7 & 91.4 & 9.1 & 90.1 \\

(h) & \cmark   &  \cmark & \cmark & \cmark &
\cmark &\cmark & 8.2 &  91.9 & 7.7 & 92.1 \\

\bottomrule

\end{tabular}%
	\label{table:ablate_kt360_1}
    }
    \\
    \begin{minipage}{\linewidth}
        \scriptsize
        \vspace{0.4em}
    \end{minipage}
\end{table}

% ----------------- Ablation Study of Heterogeneous Pinhole-Fisheye Configuration on RealHet  -----------------
\begin{table}[t]
    \centering
    \caption{
        \textbf{Ablation study}
        of the effectiveness in heterogeneous fisheye-pinhole configurations on \PrivateDatasetOne{} dataset. ``F\_0'' and ``F\_1'' refer to the front and left fisheye cameras, while ``P\_0'', ``P\_1'' and ``P\_2'' represent the front, left front and left back pinhole cameras.
    }
    \resizebox{\linewidth}{!}{
	\large 
\begin{tabular}{c @{}cc ccc c c @{}c @{}c  c @{}c @{}c}
\toprule

\multicolumn{1}{c}{\multirow{2}{*}{Exp.}} &
\multicolumn{2}{c}{Front View} &
\multicolumn{3}{c}{Left View} &
\multicolumn{1}{c}{\multirow{2}{*}{\IdeaOne{}}} &
\multicolumn{3}{c}{Fisheye Performance} &
\multicolumn{3}{c}{Pinhole Performance}
\\
&
\multicolumn{1}{c}{F\_0} &
\multicolumn{1}{c}{P\_0} &
\multicolumn{1}{c}{F\_1} &
\multicolumn{1}{c}{P\_1} &
\multicolumn{1}{c}{P\_2} &
&
AbsRel↓ & &
MSE(m)↓ &
AbsRel↓ & &
MSE(m)↓
\\

\midrule      

(a) &\cmark &  &  &  & &
\xmark & 7.1 & & 0.277 & - & & -\\

(b) & &\cmark & &  & &
\xmark & - & & - & 7.6 & & 0.582 \\ 

(c) & \cmark & \cmark & &  & &
\cmark & 6.5 & & 0.229 & 6.7 & & 0.504  \\ 

\midrule

(d) & & & \cmark  &  & &
\xmark & 5.9 & & 0.193 & - & & -\\

(e) & & & & \cmark & \cmark &
\xmark & - &  & - & 6.4 & & 0.387 \\  

(f) & & & \cmark & \cmark  & \cmark &
\cmark & 5.6 & & 0.161 & 6.1 & & 0.278 \\ 

\bottomrule

\end{tabular}%
	\label{table:ablate_realhet}
    }
    \\
    \begin{minipage}{\linewidth}
        \scriptsize
        \vspace{0.4em}
    \end{minipage}
\end{table}

% ----------------------------------
Since we are the pioneers in the novel pinhole-fisheye multi-view depth estimation configuration, we first take the only open-source pinhole-fisheye public dataset \textit{KITTI-360}. To demonstrate the generalizability, we further collect heterogeneous multi-view vehicle video data in real scenarios create an internal dataset, which are named as \PrivateDatasetOne{}. We benchmark our results on one public dataset and one internal dataset above.

\noindent \textbf{\textit{KITTI-360.}
} The \textit{KITTI-360} dataset includes two pinhole cameras in the front, and two fisheye cameras in the left and right side (See Setting 1 in \cref{fig:method_main_network}), consisting of $58345 \times 4$ images in 8 sequences. We use the entire sequence 0-7 for training and sequence 9 for testing. The original fisheye image resolution is $1400 \times 1400$ with $185 ^\circ$ FOV, while the pinhole image resolution is $376 \times 1408$. For training convenience, we uniformly resize image resolution for all views to $352 \times 640$, while intrinsic parameters undergo parameter scaling operations in accordance with image resolution.

\noindent \textbf{\PrivateDatasetOne{}.
} \PrivateDatasetOne{} is an internally collected driving dataset, where four surround-view fisheye images and six pinhole images are equipped on various vehicles (See Setting 2 in \cref{fig:method_main_network}), which is a typical camera configuration on today's commercial electric vehicles. \PrivateDatasetOne{} has collected $169$ scenarios in total with $ 60592 \times 10$ images, including indoor parking garage, outdoor urban street, apartment community, and \textit{etc} under daytime and night conditions. We randomly split $90\%$ scenarios for training and the remaining $10\%$ scenarios for testing. The training and inference resolution is set to match that of \textit{KITTI-360}.

%%%%%%%%%%%%%%%%%%%%%%%%%%%%% Evaluations %%%%%%%%%%%%%%%%%%%%%%%%%%%%%
\subsection{Evaluations}
\label{sec:evaluation}

% ----------------------------------

\noindent \textbf{Evaluation Metrics.
}
Following widely recognized research practice, we adopt Absolute Mean Relative Error (\textit{AbsRel}), Root Mean Square Error (\textit{RMSE}), and $\delta_1$ accuracy as evaluation protocols. The detailed illustration of these protocols are presented previous research work \cite{kim2022self, ke2024repurposing, yang2024depthanything_v2}.

% ----------------------------------

\noindent \textbf{Quantitative Comparison with Other Methods.
}
  We compare \method{} to widely-used monocular zero-depth estimation methods and surround-view depth estimation methods. \textbf{\method{} achieves state-of-the-art performance in most cases.} Compared to widely recognized monocular depth estimation methods, such as large discriminative model Depth Anything \cite{yang2024depth}, and large generative model Marigold \cite{ke2024repurposing}, we found that our \method{}, along with other previous multi-view methods like Surround-Depth \cite{wei2023surrounddepth}, still performs better on distorted fisheye images. It indicates that zero-shot monocular depth models still have limitations in distorted fisheye images, for which the inherent pinhole knowledge priors that is stored in such large models may pose challenge to balance new knowledge learning and old knowledge forgetting. It also suggests that multi-view network can exhibit its importance by creating 360-degree spatial information as a global context, which largely help perceive and reconstruct distorted fisheye images.

% ----------------- Qualitative Comparision on KITTI-360  -----------------
\begin{figure}[t]
    \centering
    \includegraphics[width=\linewidth]{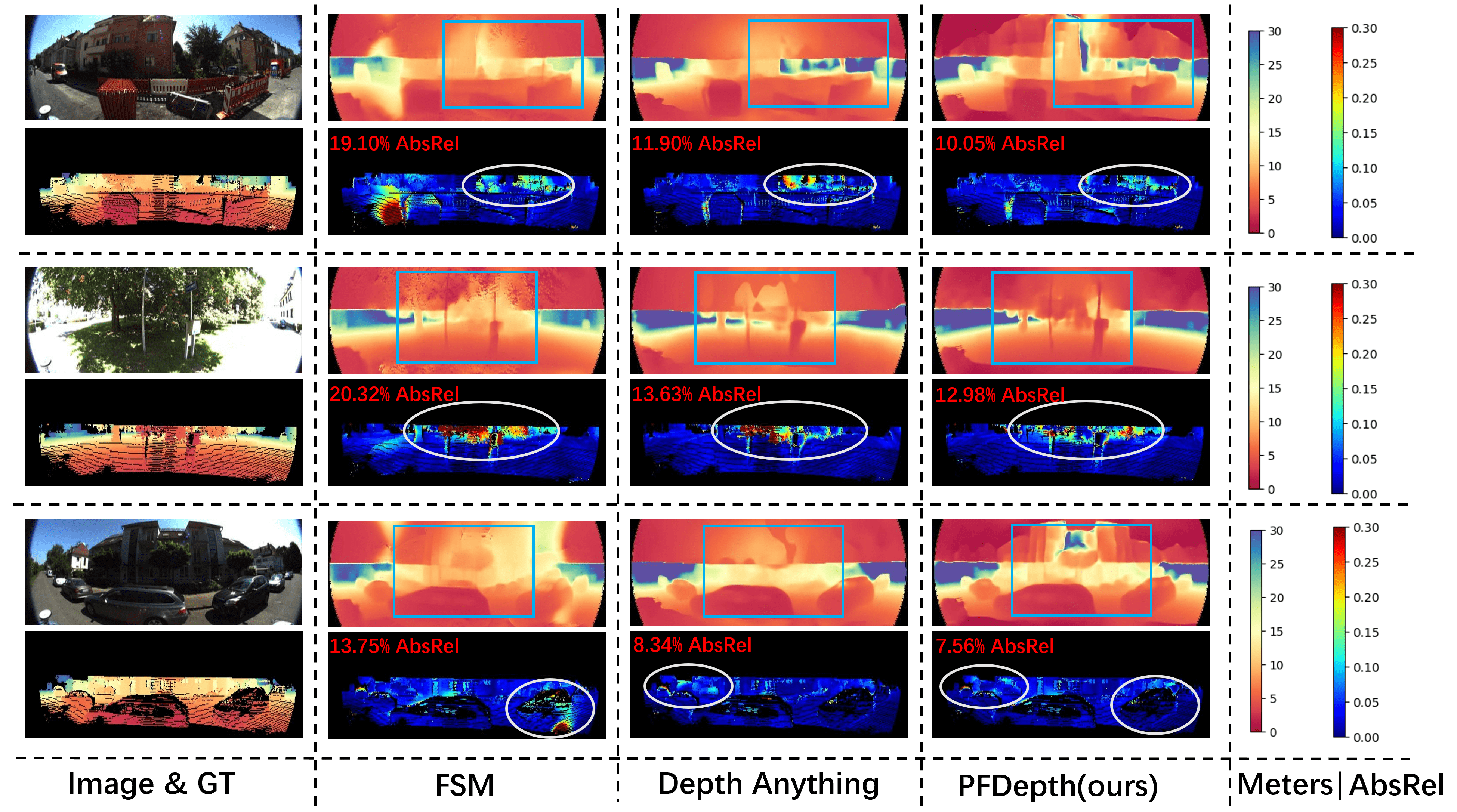}
    \vspace{-1 em}
    \caption{The visualization of fisheye predictions.
    }
    \Description{}
    \label{fig:vis_compare_fisheye}
    \vspace{-1 em}
\end{figure}

\begin{figure}[t]
    \centering
    \includegraphics[width=\linewidth]{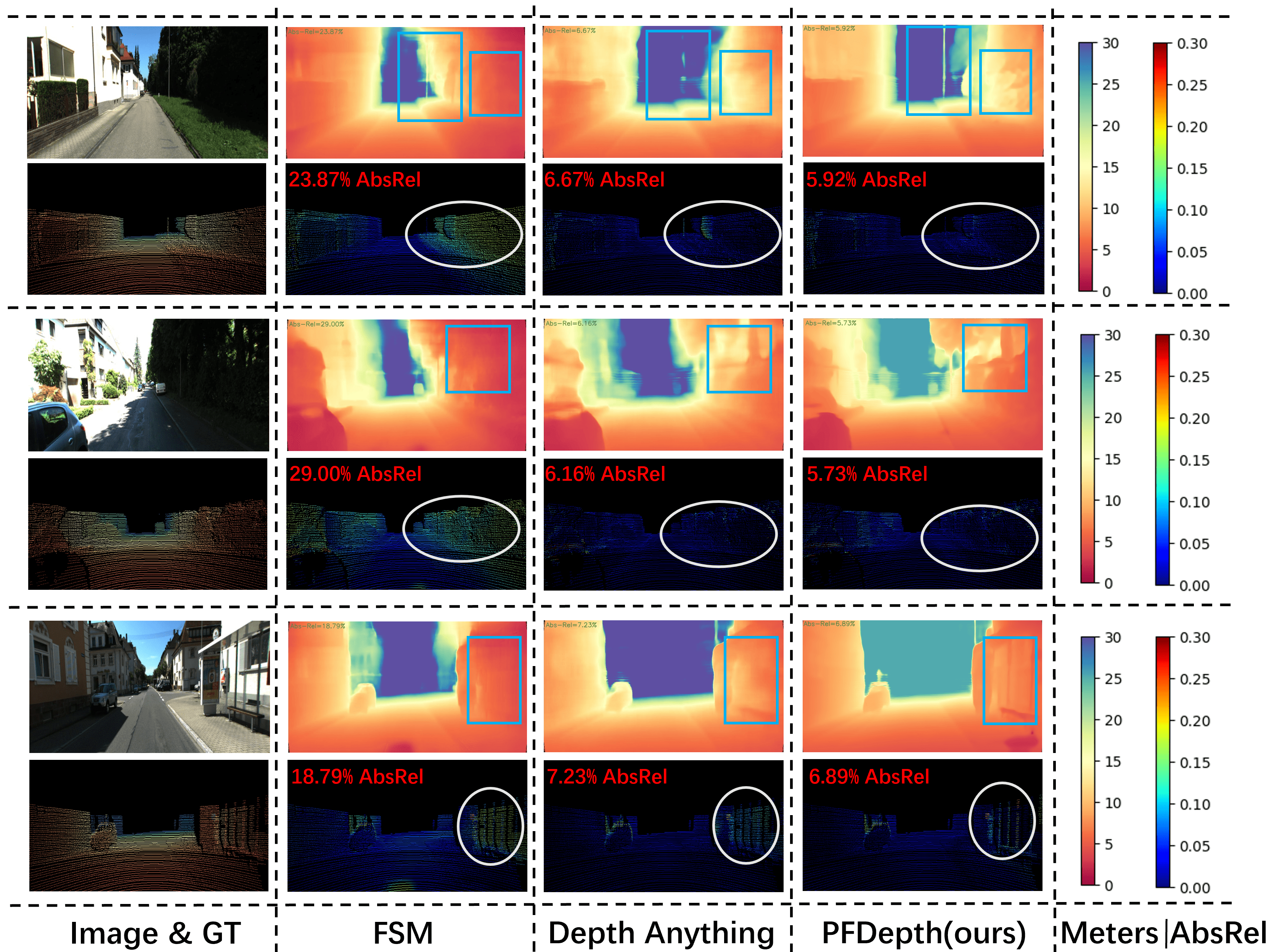}
    \vspace{-1 em}
    \caption{The visualization of pinhole predictions.
    }
    \Description{}
    \label{fig:vis_compare_pinhole}
\end{figure}

% --- fig:hsf_ablate
\begin{figure}[t]
    \centering
    \includegraphics[width=0.48\textwidth]{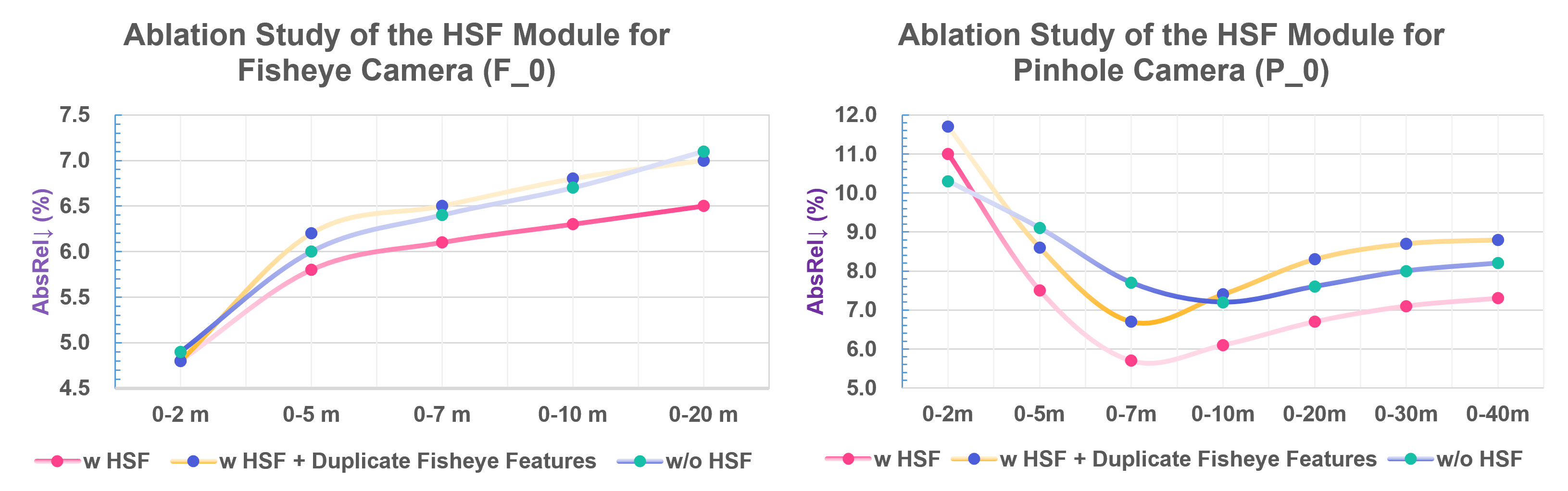}
    \vspace{-1em}
    \caption{Ablation study of the HSF module.
    }
    \Description{}
    \label{fig:exp_hsf_ablate}
    \vspace{-1em}
\end{figure}

% ----------------------------------

\noindent \textbf{Qualitative Comparison with Other Methods.
}
We compare our fisheye new results in \cref{fig:teaser} and  \cref{fig:vis_compare_fisheye}, while also evaluating traditional stereo pinhole visualization in \cref{fig:vis_compare_pinhole} and \cref{fig:teaser}. It is surprising that our visualization results on both domains perform well, where detailed textures of background hash are exhibited well without over-fitting too much (maintaining a relatively low ABS-Rel value). It should be noted that the ground-truths of \textit{KITTI-360} have a very thin distribution, leaving the ground and the sky area uncovered. This will result in almost no punishment from supervised training via SILog and L1 loss, which ultimately leads to the wide appearance of very close values (in red) distributed in "no GT areas" among almost every method. However, our methods with HSF and 3DGS can try their best to estimate depth values in such areas without a performance drop, which can demonstrate our robustness and generalization over unlabeled areas.

%%%%%%%%%%%%%%%%%%%%%%%%%%%%% Ablation Study %%%%%%%%%%%%%%%%%%%%%%%%%%%%%
\subsection{Ablation Study}
\label{sec:ablation}
To dig deeper into the effectiveness of our heterogeneous setting and spatial aggregation module, we conduct extensive ablation study in various aspects, where we found much insightful observations.

% ----------------------------------

\noindent \textbf{Ablation of the Heterogeneous Configuration.
}
To validate the necessity of our heterogeneous pinhole-fisheye configurations, we conduct an extensive ablation study on \textit{KITTI-360} and \PrivateDatasetOne{} datasets (See \cref{table:ablate_kt360_1} and \cref{table:ablate_realhet}).
From comprehensive ablation on various heterogeneous settings, it can be fully observed that: (1) directly mixing pinhole and fisheye images for training dramatically degrades performance, especially for pinhole images; (2) conversely, the HSF module effectively ensures better training for heterogeneous configurations compared to directly mixing views, and especially improves fisheye performance at a consistent scale;  (3) in \textit{KITTI-360}, pinhole performance under heterogeneous setting slightly drops compared to pure pinhole baseline (when w/o. 3DGS but with HSF); but in \PrivateDatasetOne{}, pinhole dataset can boost consistent improvement. The possible factor may be attributed to the large discrepancy in aspect ratio of pinhole images ($376 \times 1408$) and fisheye images ($1400 \times 1400$) in \textit{KITTI-360}, which are uniformly resized to $352 \times 640$ for convenient implementation. Yet in current commercial electrical vehicles, the aspect ratio of fisheye and pinhole images is the same or similar, like \PrivateDatasetOne{}.

% ----------------------------------

\noindent \textbf{Ablation of the Heterogeneous Spatial Fusion module.
}
 To further explain the internal principle, we further ablate the Heterogeneous Spatial Fusion (HSF) module between the front pinhole camera and fisheye camera on \PrivateDatasetOne{} dataset, as shown in \cref{fig:exp_hsf_ablate}. In addition to the setting with or without HSF, we add a new comparative setting (see \cref{fig:exp_hsf_ablate}) where we substitute the overlapped pinhole voxel features in $\mathbf{V}_o$ with duplicated fisheye voxel features for fisheye evaluation, and keep the HSF network unchanged to control the unique variable. We also substitute the overlapped fisheye voxel features with duplicated pinhole features for pinhole evaluation. An interesting phenomenon is observed that the combination of the overlapped pinhole and fisheye features matters most in the improvement of performance, because the lack of pinhole voxel features will drop fisheye performance in distant areas more, while the lack of fisheye voxel features will diminish pinhole performance in middle and distant ranges as well. 

% ----------------------------------

%% Conclusion
\section{Conclusion}
\label{sec:conclusion}
We propose a brand-new heterogeneous multi-view perception framework \method{}, including distortion-aware feature extraction, Heterogeneous Spatial Fusion (\IdeaOne{}) module, and a Gaussian-Splatted dynamic fusion method across heterogeneous views in 3D space. Our results exhibit promising improvement by leveraging the complementary benefits of the pinhole-fisheye imagery system.

%% Acknowledgement
\begin{acks}
The research is supported by
 National Natural Science Foundation of China
 (No.72192821, No.62472282), The Fundamental Research Funds for the Central Universities (project number: YG2023QNA35), Natural Science Foundation of Shanghai (25ZR1402135), and Shanghai Key Laboratory of Computer Software Evaluating and Testing.
\end{acks}

\bibliographystyle{ACM-Reference-Format}
\bibliography{full_paper_ref}

\appendix

\section{Algorithm descriptions of Heterogeneous Spatial Fusion}
\label{sec:appendix_HSF_algorithm}

In \cref{sec:msa} we introduce the Heterogeneous Spatial Fusion (HSF) module, which involves two principal stages:
1) Identification of overlapped and non-overlapped regions, and 2) Fusion of these regions across multiple views. The following two algorithms \cref{alg:overlap_mask} and \cref{alg:hsf} describe the procedure in detail. Algorithm~\ref{alg:overlap_mask} determines whether the projected position of a voxel falls on more than one image plane; a voxel is deemed to overlap views $i$ and $j$ when both $\mathbf{Mask}[i]$ and $\mathbf{Mask}[j]$ are \textit{true}. Leveraging the resulting masks, Algorithm~\ref{alg:hsf} separates overlapped and non-overlapped voxels, integrates them across neighbouring views into $\mathbf{V}'{O_i}$ and $\mathbf{V}'{N_i}$, respectively, and finally consolidates these into the global representations $\mathbf{V}{O}$ and $\mathbf{V}{N}$, which are concatenated for subsequent processing.

\setcounter{AlgoLine}{0}
\begin{algorithm}[htbp!]
    \KwIn{Predefined voxel center tensors $\mathbf{S}\in \mathbb{R}^{N \times 4}$ at ego-vehicle coordinate, where $N=XYZ$ is the number of voxels; $m$ pinhole views and $n$ fisheye views with corresponding camera parameters $H_{\{i, j\}}, W_{\{i, j\}}, \mathbf{K}_{\{i, j\}}, \mathbf{E}_{\{i, j\}}, \mathbf{\omega}_j$.}
    \KwOut{Binary mask tensors $\mathbf{Mask}\in \mathbb{R}^{(m+n) \times N \times 1}$}
    \BlankLine
    \caption{Calculation of Overlapped and Non-overlapped Voxels}
    \label{alg:overlap_mask}
    
    \For{$i \leftarrow 1$ to $m$}{
        $\mathbf{S} \leftarrow \mathbf{S}$.transpose() $\quad$ \tcp{$\mathbf{S} \in \mathbb{R}^{N \times 4} \rightarrow \mathbb{R}^{ 4 \times N} $}
        
        $\mathbf{S}^{'} \leftarrow \Pi_P (\mathbf{S}, \mathbf{E}_i, \mathbf{K}_i)$  $\quad$ \tcp{ $\mathbf{S}^{'} \in 
 \mathbb{R}^{3 \times N} $}
 
        $\mathbf{Mask}[i]$ = ($0 \le \mathbf{S}^{'}[0] \le H_i$) AND ($0 \le \mathbf{S}^{'}[1] \le W_i$)
    }

    \For{$j \leftarrow m+1$ to $m+n$}{
        $\mathbf{S} \leftarrow \mathbf{S}$.transpose()
        
        $\mathbf{S}^{''} \leftarrow \Pi_F (\mathbf{S}, \mathbf{E}_j, \mathbf{K}_j, \mathbf{\omega}_j)$
        
        $\mathbf{Mask}[j]$ = ($0 \le \mathbf{S}^{''}[0] \le H_j$) AND ($0 \le \mathbf{S}^{''}[1] \le W_j$)
    }

    \Return($\mathbf{Mask}$)

\end{algorithm}
\setcounter{AlgoLine}{0}
\begin{algorithm}[htbp!]
    \KwIn{View-specific voxel features $\mathbf{V}_i \in \mathbb{R}^{X \times Y \times Z \times C}$; binary mask tensors $\mathbf{Mask} \in \mathbb{R}^{(m+n) \times N \times 1}$, where $N=XYZ$.}
    \KwOut{Mixed voxel features $\mathbf{V}_{\text{mixed}} \in \mathbb{R}^{X \times Y \times Z \times C}$}
    \BlankLine
    \caption{Heterogeneous Spatial Fusion Process}
    \label{alg:hsf}

    Create zero tensor $\mathbf{V}_O, \mathbf{V}_N \in \mathbf{R}^{X \times Y \times Z \times C}$ 
    
    \For{$i \leftarrow 1$ to $m+n-1$}{
        $\mathbf{M}_{O_i}$ $\leftarrow$  $\mathbf{Mask}[i]$ AND $\mathbf{Mask}[i+1]$
        
        $\mathbf{M}_{N_i} \leftarrow$ NOT $\mathbf{M}_{O_i}$
        
        $\mathbf{V}_{i}^{'} \leftarrow \mathbf{V}_{i} + \mathbf{V}_{i+1}$
        
        $\mathbf{V}_{i}^{'} \leftarrow \mathbf{V}_{i}^{'}$.reshape$(N, C)$

        \tcc{Process overlapped voxels}
    
        $\mathbf{V}_{O_i} \leftarrow$ ($ \mathbf{V}_{i}^{'}[\mathbf{M}_{N_i}] \leftarrow $ zero-vector)

        $\mathbf{V}_O \leftarrow \mathbf{V}_O + $ MLP($\mathbf{V}_{O_i}$).reshape($X, Y, Z, C$)

        \tcc{Process non-overlapped voxels}

        $\mathbf{V}_{N_i} \leftarrow$ ($ \mathbf{V}_{i}^{'}[\mathbf{M}_{O_i}] \leftarrow $ zero-vector)

        $\mathbf{V}_N \leftarrow \mathbf{V}_N + $ MLP($\mathbf{V}_{N_i}$).reshape($X, Y, Z, C$)
    }

    $\mathbf{V}_{\text{mixed}} \leftarrow$ Conv3D(Concat($\mathbf{V}_O$, $\mathbf{V}_N$))

    \Return($\mathbf{V}_{\text{mixed}}$)

\end{algorithm}

\section{Loss Function}
\label{sec:appendix_loss}

\noindent In \cref{sec:training_loss} we present our training objective, which combines supervised, self-supervised, and heterogeneous self-supervised terms. The overall loss is

\begin{align}
\label{formula:total_loss}
\mathcal{L}_\text{total} &= \mathcal{L}_\text{supervised}
                         + \mathcal{L}_\text{self-supervised}
                         + \mathcal{L}_\text{het-self-sup}, \\[2pt]
\mathcal{L}_\text{supervised} &= \alpha_1 \mathcal{L}_\text{L1}
                               + \alpha_2 \mathcal{L}_\text{SIlog}
                               + \alpha_3 \mathcal{L}_\text{pseu-ranking}, \\[2pt]
\mathcal{L}_\text{self-supervised} &= \alpha_4 \mathcal{L}_\text{smooth}, \\[2pt]
\mathcal{L}_\text{het-self-sup} &= \alpha_5 \mathcal{L}_\text{temporal}
                                 + \alpha_6 \mathcal{L}_\text{pf-spatial}.
\end{align}

\noindent Throughout all experiments, we fix
$\alpha_1 = 1$, $\alpha_2 = 0.1$, $\alpha_3 = 0.001$,
$\alpha_4 = 0.001$, $\alpha_5 = 0.1$, and $\alpha_6 = 0.03$.

\subsection{Supervised Loss}

We combine an $\ell_1$ loss, an SILog loss~\cite{eigen2014depth}, and a pseudo-ranking loss~\cite{chen2016single,xian2020structure}.
Let $d^{\text{pred}}$ and $d^{\text{gt}}$ denote the predicted and ground-truth disparities, and let $i \in HW$ index image pixels. The $\ell_1$ term is
\begin{align}
\label{formula:L1_loss}
\mathcal{L}_\text{L1} = \bigl\lVert d_i^{\text{pred}} - d_i^{\text{gt}}\bigr\rVert_1.
\end{align}

\noindent The scale-invariant logarithmic (SILog) loss~\cite{eigen2014depth} reduces scale ambiguity:
\begin{align}
\label{formula:SILog_loss}
r_i &= \log d_i^{\text{pred}} - \log d_i^{\text{gt}}, \\[2pt]
\mathcal{L}_\text{SIlog} &=
      \frac{1}{|HW|}\sum_{i\in HW} r_i^{\,2}
      - \bigl(\tfrac{1}{|HW|}\sum_{i\in HW} r_i\bigr)^{\!2}.
\end{align}

\noindent The pseudo-ranking loss enforces correct ordinal relations between pixel pairs:
\begin{align}
\mathcal{L}_\text{pseu-ranking} = \sum_{k=1}^{K}
      \psi_k(I, i_k, j_k, d^{\text{pred}}),
\end{align}
where the label for pair $(i_k,j_k)$ is
\begin{align}
r_k =
\begin{cases}
+1, & d_{i_k}^{\text{gt}} - d_{j_k}^{\text{gt}} >  \epsilon,\\
-1, & d_{i_k}^{\text{gt}} - d_{j_k}^{\text{gt}} < -\epsilon,\\
 0, & |d_{i_k}^{\text{gt}} - d_{j_k}^{\text{gt}}| \le \epsilon,
\end{cases}
\end{align}
and the pairwise loss is
\begin{align}
\psi_k(\!\cdot\!) =
\begin{cases}
\log\!\bigl(1 + e^{-d_{i_k}^{\text{pred}} + d_{j_k}^{\text{pred}}}\bigr), & r_k = +1,\\[4pt]
\log\!\bigl(1 + e^{\phantom{-}d_{i_k}^{\text{pred}} - d_{j_k}^{\text{pred}}}\bigr), & r_k = -1,\\[4pt]
\bigl(d_{i_k}^{\text{pred}} - d_{j_k}^{\text{pred}}\bigr)^{\!2},           & r_k = 0.
\end{cases}
\end{align}

\noindent In practice, using pseudo rankings derived from Depth-Anything~v2~\cite{yang2024depthanything_v2}
yields a modest $\!\approx\!0.1$–$0.2\%$ reduction in AbsRel, whereas sparse metric depth (``real'' rankings) offers no meaningful gain.

\subsection{Self-Supervised Loss}

We adopt an edge-aware smoothness term~\cite{Xian_2020,kim2022self}:
\begin{align}
\label{formula:smoothness_loss}
\mathcal{L}_\text{smooth} = 
\sum_{i\in HW}\!\sum_{k\in\{x,y\}}
\bigl\lVert \nabla_k d_i^{\text{pred}}\,
           e^{-\lVert \nabla_k \mathbf{I}_i\rVert}\bigr\rVert.
\end{align}

\subsection{Heterogeneous Self-Supervised Loss}

Because the ego-vehicle moves, we can self-supervise via spatial and temporal warping.  Let
$\xi(\mathbf{I},\hat{\Pi})$ denote bilinear sampling of image~$\mathbf{I}$ using 2-D coordinates produced by transformation~$\hat{\Pi}$.

\paragraph{Spatial warping.}  For view $i\!\rightarrow\!j$ we define
\begin{align}
\label{formula:appendix_sp_warp}
W_{i \rightarrow j}^{t} = 
\begin{cases}
  \xi(\mathbf{I}_{F_j},\Pi_{F_j}\Pi_{F_i}^{-1}), & F_i \!\rightarrow\! F_j,\\
  \xi(\mathbf{I}_{P_j},\Pi_{P_j}\Pi_{F_i}^{-1}), & F_i \!\rightarrow\! P_j,\\
  \xi(\mathbf{I}_{F_j},\Pi_{F_j}\Pi_{P_i}^{-1}), & P_i \!\rightarrow\! F_j,\\
  \xi(\mathbf{I}_{P_j},\Pi_{P_j}\Pi_{P_i}^{-1}), & P_i \!\rightarrow\! P_j,
\end{cases}
\end{align}
where $\Pi_{\{\cdot\}}$ and $\Pi_{\{\cdot\}}^{-1}$ are the projection and un-projection operators of the corresponding (pinhole~$P$ or fisheye~$F$) camera.

\paragraph{Temporal warping.}
For successive frames $t\!\rightarrow\!t'$ of camera~$i$ we warp via
\begin{align}
\label{formula:appendix_temp_warp}
W_i^{t \rightarrow t'} =
\xi\!\bigl(\mathbf{I}_{\{P,F\}},\,\Pi_{\{P_i,F_i\}}\,
          \mathbf{\hat{T}}^{t \rightarrow t'}\,
          \Pi_{\{P_i,F_i\}}^{-1}\bigr),
\end{align}
where $\mathbf{\hat{T}}^{t \rightarrow t'}$ is the estimated ego-motion.

\paragraph{Photometric losses.}  With $\beta\!=\!0.85$ we compute
\begin{align}
\label{formula:appendix_ssim_loss}
\mathcal{L}_\text{pf-spatial} &=
(1-\beta)\lVert \hat{\mathbf{I}}_j - \mathbf{I}_i\rVert_1
+ \beta\,\frac{1 - \text{SSIM}(\hat{\mathbf{I}}_j,\mathbf{I}_i)}{2},\\[4pt]
\mathcal{L}_\text{temporal} &=
(1-\beta)\lVert \hat{\mathbf{I}}_i^{t'} - \mathbf{I}_i^{t}\rVert_1
+ \beta\,\frac{1 - \text{SSIM}(\hat{\mathbf{I}}_i^{t'},\mathbf{I}_i^{t})}{2}.
\end{align}

\subsection{Ablation Study}

\begin{figure}[htbp!]
    \centering
    \includegraphics[width=1.0\linewidth]{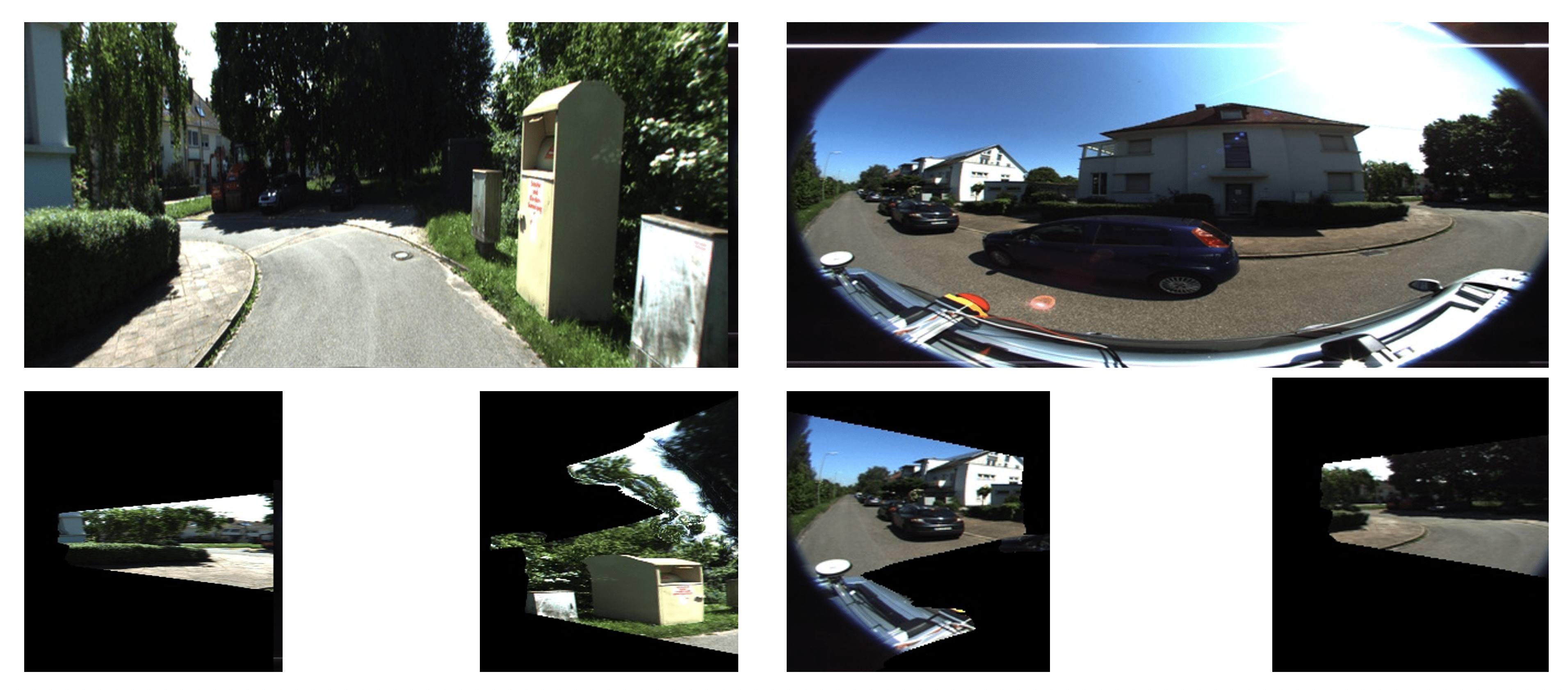}
    \caption{Overview of heterogeneous spatial warp visualization. }
    \Description{}
    \label{fig:appendix_spatial_warp_vis}
\end{figure}

\begin{figure}[htbp!]
    \centering
    \includegraphics[width=1.0\linewidth]{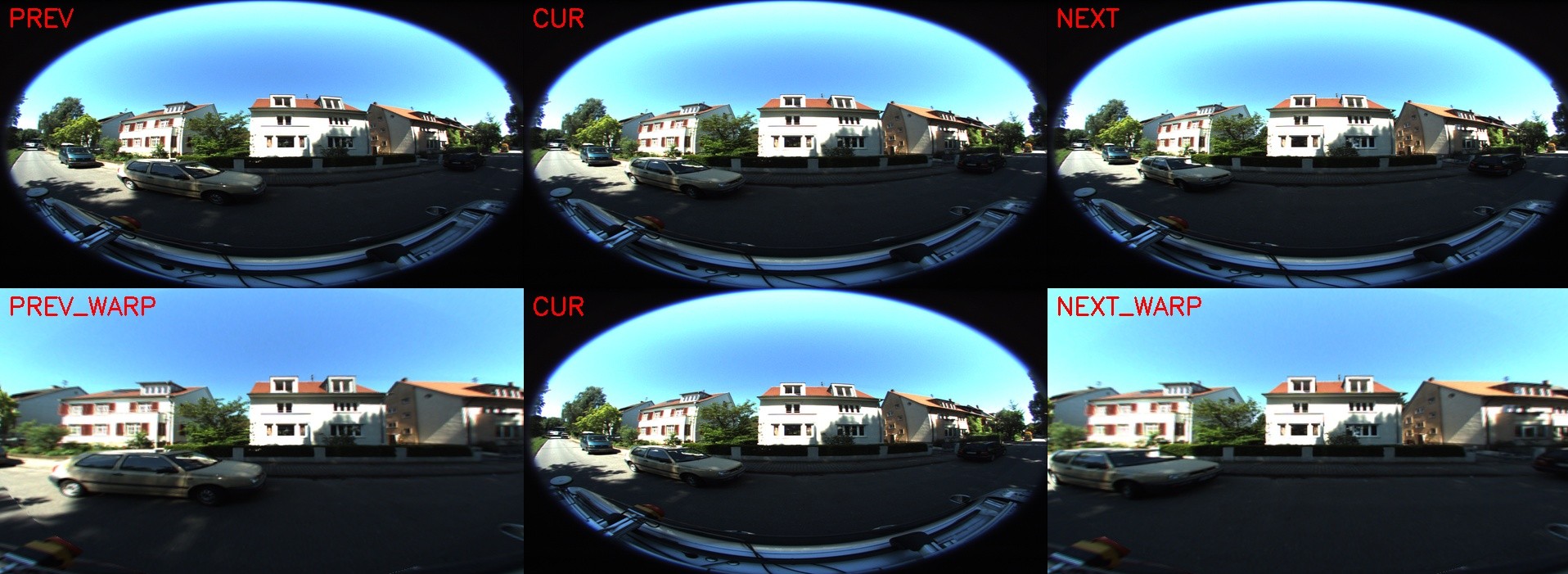}
    \caption{ Overview of heterogeneous temporal warp visualization.}
    \Description{}
    \label{fig:appendix_temp_warp_vis}
\end{figure}

Quantitative and qualitative ablations are reported in
\cref{tab:appendix_loss_ablation},
\cref{fig:appendix_spatial_warp_vis},
and \cref{fig:appendix_temp_warp_vis}.
Either $\mathcal{L}_\text{L1}$ or $\mathcal{L}_\text{SIlog}$ alone almost matches the full-loss performance; omitting \emph{both} degrades results markedly.
Adding the self-supervised terms leaves AbsRel nearly unchanged ($<\!0.1\%$) yet visibly sharpens boundaries and fine textures.
The warped views in \cref{fig:appendix_spatial_warp_vis} and
\cref{fig:appendix_temp_warp_vis} closely match their targets,
demonstrating the stability and accuracy of the estimated depths.

% --- Appendix HSF, 3DGS Ablation Figure

\begin{table}[htbp]
  \centering
  \caption{Ablation study of different loss configurations.}
  \label{tab:appendix_loss_ablation}
  \begin{tabular}{l c}
    \toprule
    Loss type                    & AbsRel $\downarrow$ [\%] \\
    \midrule
    Full                         & $+0$   \\
    w/o L1 loss                  & $+0.2$ \\
    w/o SILog loss               & $+0.3$ \\
    w/o SILog nor L1 loss        & $+7.0$ \\
    w/o pseu-ranking loss        & $+0.1$ \\
    \bottomrule
  \end{tabular}
\end{table}

% ============ Supplementary Visualisations ============

\section{Supplementary Visualisation Results}
\label{sec:supp_vis}

% --- HSF / 3DGS Ablation Figure
\begin{figure*}[htbp]
    \centering
    \includegraphics[width=\linewidth]{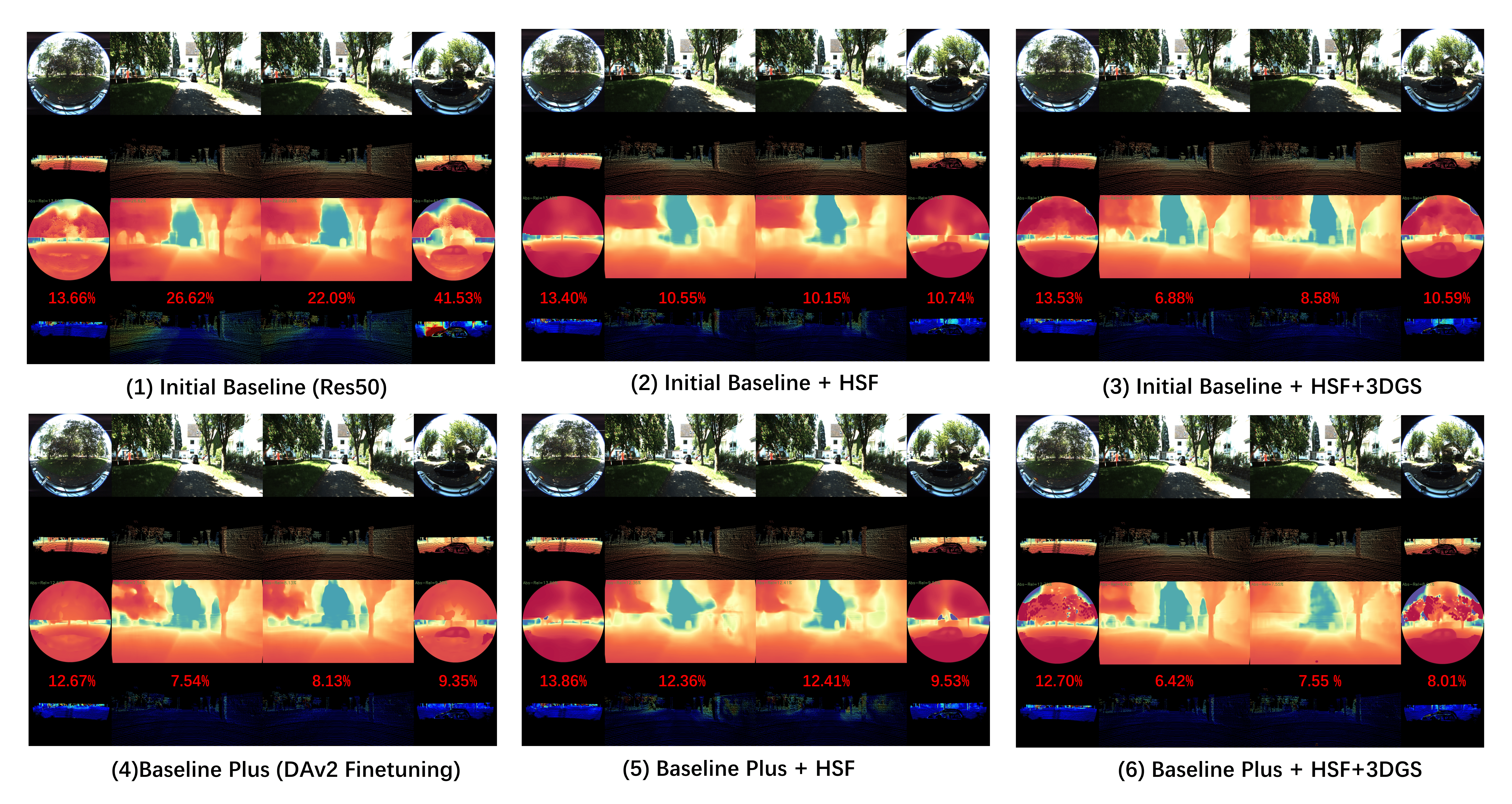}
    \caption{Qualitative ablation of HSF and 3DGS on different backbones (best viewed zoomed in).
             ``DAv2'' denotes a Depth-Anything~v2 model fine-tuned on KITTI-360 metric LiDAR depth.}
    \label{fig:appendix_visual_ablation}
\end{figure*}

% --- Surround-view Example Figure
\begin{figure*}[htbp]
    \centering
    \includegraphics[width=0.6\linewidth]{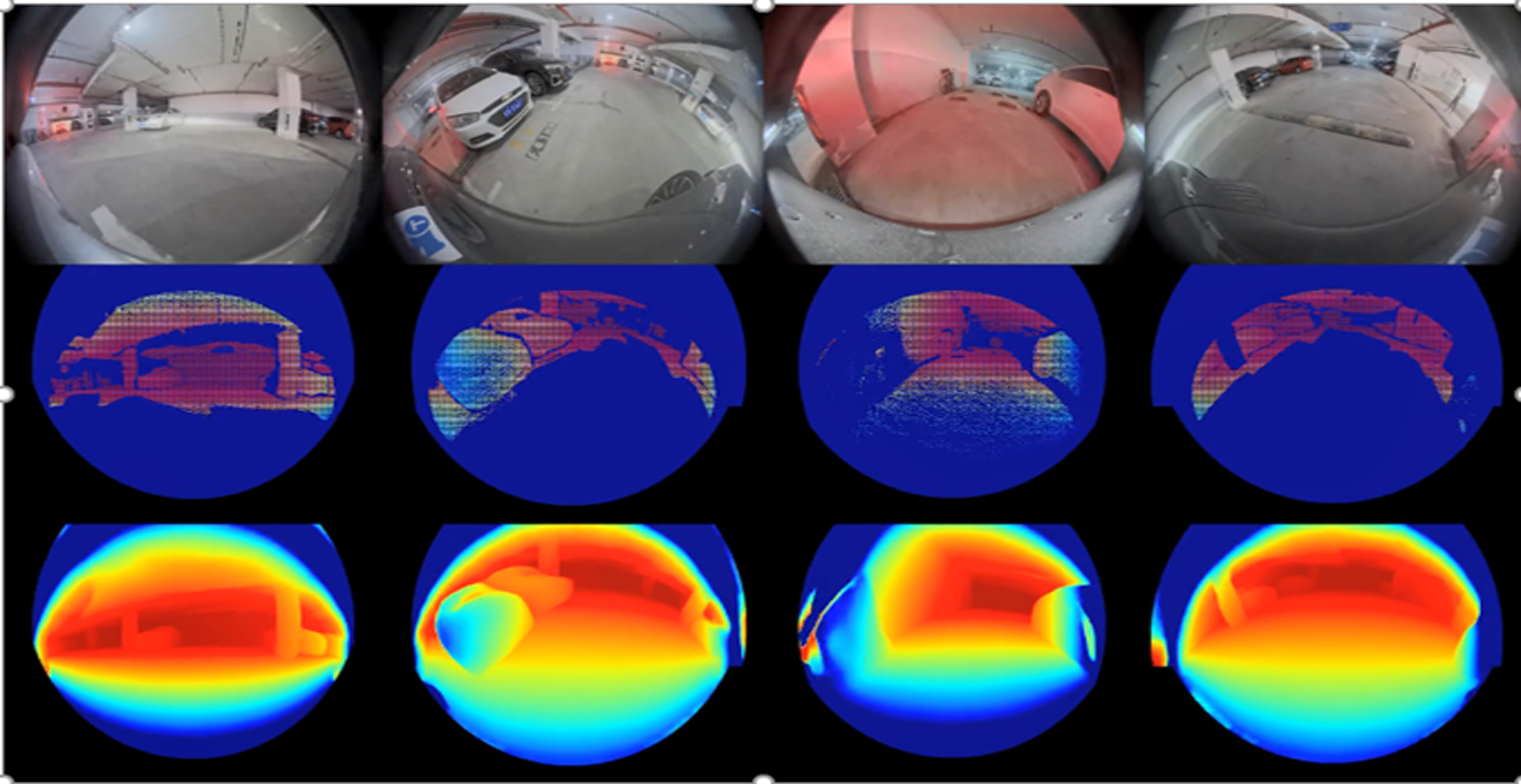}\hfill
    \includegraphics[width=0.7\linewidth]{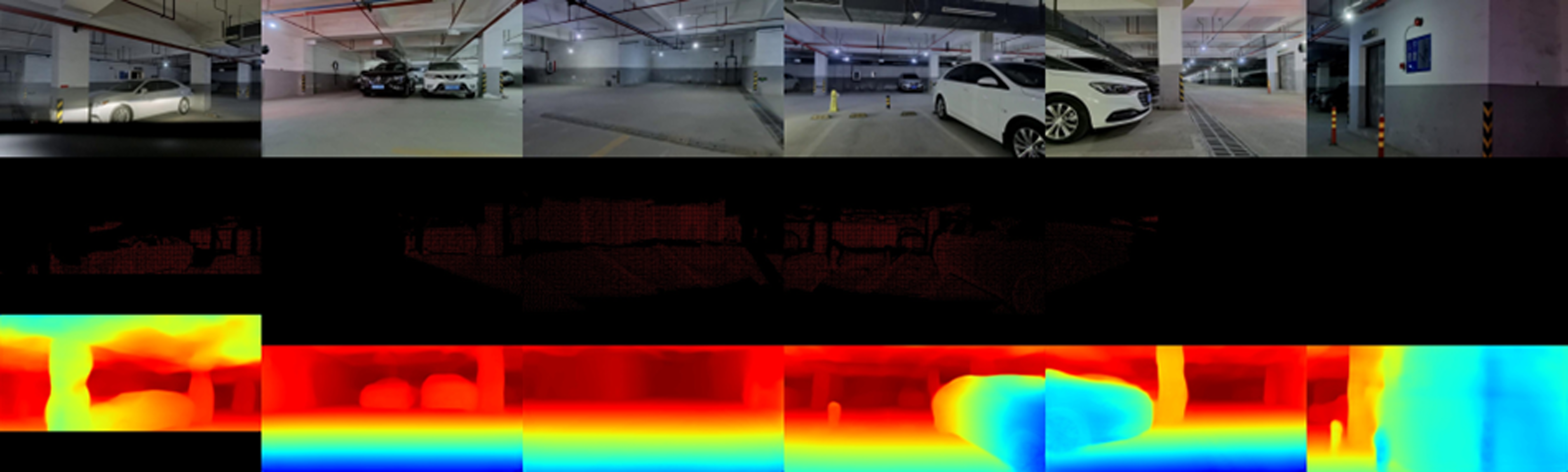}
    \caption{Surround-view predictions (four fisheye + six pinhole) on \PrivateDatasetOne{}.}
    \label{fig:appendix_realhet_fisheye}
\end{figure*}

\subsection{Qualitative Ablation of HSF and 3DGS}
\label{sec:qual_ablation}

Figure~\ref{fig:appendix_visual_ablation} compares six model variants on fisheye–pinhole inputs and reveals several noteworthy trends:

\begin{itemize}
\item \textbf{Setting (1) vs.\ Setting (2).}  
      The ResNet-50 encoder–decoder (setting~1) produces visually rich 2D textures but exhibits poor depth accuracy.  
      Introducing HSF (setting~2) smooths low-level details yet substantially improves spatial coherence, indicating that the bare 2-D architecture lacks explicit 3-D awareness.

\item \textbf{Setting (2) vs.\ Setting (3).}  
      Although HSF enhances geometric reasoning, it also suppresses certain semantic priors.  
      Adding dynamic 3DGS sampling (setting~3) restores fine-grained structure—especially for thin objects, foliage, low-light regions, and LiDAR-sparse zones—while further boosting quantitative metrics.  
      Given its negligible overhead (see Table~\ref{table:quan_main}), 3DGS acts as an effective spatial-awareness prompt.

\item \textbf{Setting (1) vs.\ Setting (4).}  
      It can be observed that with larger backbone and empowered scales of data for pre-training (setting~4, DAv2), pure 2D encoder-decoder model can gradually mitigate ``texture overfitting'' observed in setting~1, and largely improve spatial-awareness. However, such improvements demand significantly more computation and data, whereas our lightweight HSF+3DGS design achieves comparable gains with a small network.

\item \textbf{ Setting (4) vs.\ Setting (5) and Setting (6).}  
      Directly inserting HSF into a large 2-D foundation model (setting~5) can hamper adaptation, suggesting that heavily pre-trained networks possess stronger inertia.  
      By contrast, augmenting the same model with both HSF and 3DGS (setting~6) recovers and surpasses its baseline accuracy.  
      These results imply that 3DGS serves as an efficient bridge between 2-D priors and 3-D geometry, easing the transition toward spatially-aware vision models.
\end{itemize}

\subsection{Limitations}
Our heterogeneous framework currently requires calibrated camera intrinsics and extrinsics, restricting its use on large-scale, uncalibrated Internet imagery.  
Moreover, richly annotated fisheye data remain scarce compared to monocular pinhole datasets, limiting opportunities for massive-scale pre-training.  
Our future work will explore combining zero-shot monocular relative-depth models with specialised small modules such as our \method{}, enabling mutual improvement via parameter-efficient fine-tuning.

\subsection{Additional Visual Examples}
\cref{fig:appendix_realhet_fisheye} presents further six pinhole-four fisheye surround-view results on \PrivateDatasetOne{}.  
Additional examples and a demo viewer will be released on the project website.

\end{document}